\documentclass[default,iicol]{sn-jnl}% Default with double 

\usepackage{graphicx}%
\usepackage{multirow}%
\usepackage{amsmath,amssymb,amsfonts}%
\usepackage{amsthm}%
\usepackage{mathrsfs}%
\usepackage[title]{appendix}%
\usepackage[dvipsnames]{xcolor}
\usepackage{textcomp}%
\usepackage{manyfoot}%
\usepackage{booktabs}%
\usepackage{algpseudocode}%
\usepackage{listings}%
\usepackage{array}
\usepackage{dsfont}

\theoremstyle{thmstyleone}

\theoremstyle{thmstyletwo}

\theoremstyle{thmstylethree}% 

\usepackage[ruled]{algorithm2e}
\begin{document}

\title[]{A Comprehensive Review of Few-shot Action Recognition}

\author[1,2]{\fnm{Yuyang} \sur{Wanyan}}\email{wanyanyuyang2021@ia.ac.cn}
\author[1,2,3]{\fnm{Xiaoshan} \sur{Yang}}\email{xiaoshan.yang@nlpr.ia.ac.cn}
\author[1,2]{\fnm{Weiming} \sur{Dong}}\email{weiming.dong@ia.ac.cn}
\author[1,2,3]{\fnm{Changsheng} \sur{Xu}}\email{csxu@nlpr.ia.ac.cn}
\affil{\textsuperscript{1}State Key Laboratory of Multimodal Artificial Intelligence System, Institute of Automation, Chinese Academy of Sciences, 100190, Beijing, China}
\affil{\textsuperscript{2}School of Artificial Intelligence, University of Chinese Academy of Sciences, 101408, Beijing, China}
\affil{\textsuperscript{3}Peng Cheng Laboratory, 518066, Shenzhen, China}

\abstract{
Few-shot action recognition aims to address the high cost and impracticality of manually labeling complex and variable video data in action recognition. 
It requires accurately classifying human actions in videos using only a few labeled examples per class. 
Compared to few-shot learning in image scenarios, few-shot action recognition is more challenging due to the intrinsic complexity of video data. 
Numerous approaches have driven significant advancements in few-shot action recognition, which underscores the need for a comprehensive survey. 
Unlike early surveys that focus on few-shot image or text classification, we deeply consider the unique challenges of few-shot action recognition. 
{\color{black}
In this survey, we provide a comprehensive review of recent methods and introduce a novel and systematic taxonomy of existing approaches, accompanied by a detailed analysis.
We categorize the methods into generative-based and meta-learning frameworks, and further elaborate on the methods within the meta-learning framework, covering aspects: video instance representation, category prototype learning, and generalized video alignment. 
}
Additionally, the survey presents the commonly used benchmarks and discusses relevant advanced topics and promising future directions. 
We hope this survey can serve as a valuable resource for researchers, offering essential guidance to newcomers and stimulating seasoned researchers with fresh insights. 
}

\keywords{Few-shot learning, action recognition, meta-learning, video classification, survey}

\maketitle

\section{Introduction}

\begin{figure*}[ht]
    \centering
    \includegraphics[width=\linewidth]{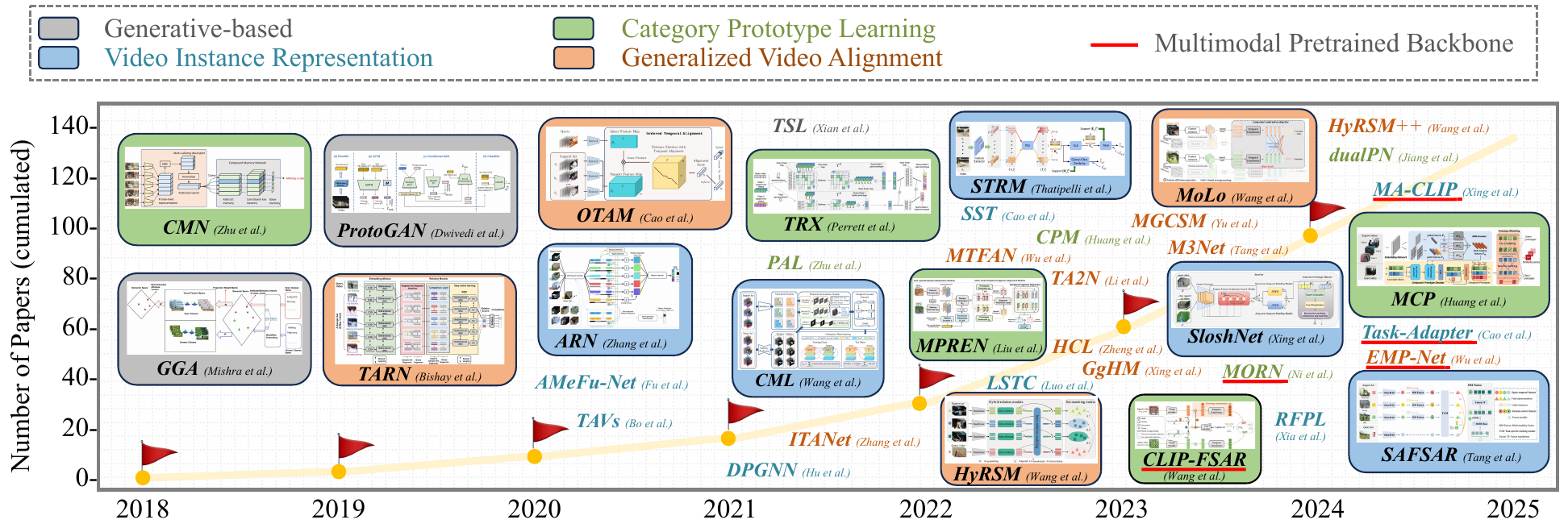}
    \caption{\color{black}
    A taxonomically organized chronological timeline of few-shot action recognition methodologies, highlighting key developments and emerging trends. 
    The plot illustrates the field's growth by showcasing the cumulative temporal distribution of publications from 2018 to 2025.
    The {\color{red}\underline{red underline}} signifies the method employs a language–image pretraining backbone (e.g., CLIP), while others adopt a unimodal pretraining backbone (e.g., ResNet). 
    }
    \label{fig:timeline}
\end{figure*}

With the rapid advancements and innovations in communication and media creation technologies, video has increasingly emerged as a prominent form of information dissemination. 
Recognizing and understanding human actions from videos is a critical topic for numerous real-world applications.
The emergence of deep learning has significantly advanced the frontiers of action recognition~\cite{wang2016temporal, wang2021tdn}. 
However, traditional deep learning models typically require a large amount of labeled video data to achieve optimal performance. 
This hampers the widespread application of deep learning techniques in many real-world scenarios, where obtaining and annotating extensive video datasets is both costly and time-prohibitive. 
Few-shot action recognition (FSAR)~\cite{zhu2018CMN} has been proposed to address this challenge. 
FSAR aims to enable models to learn and generalize effectively with only a few labeled videos, thereby reducing data dependency and enhancing application efficiency.

In contrast to few-shot learning in image or text scenarios~\cite{qiao2018few, cai2018memory, schick2021exploiting}, few-shot action recognition presents unique challenges because videos encompass both visual information within frames and temporal dynamics between frames. 
Recognizing human actions necessitates the careful modeling of temporal sequences and rich semantic information, beyond merely identifying human appearance, objects, and the context in each frame. 
For instance, the action of ``underwater diving" requires not only identifying water environments and divers but also understanding the trajectories to distinguish fine-grained diving actions such as ``ascent", ``balance", and ``descent", which are temporally distinct despite their similar appearances. 
This poses a significant challenge in extracting effective features for human action recognition. 
Consequently, learning intricate video representations that can be generalized to novel categories remains a difficult task. 
Moreover, the issue of intra-class variance becomes increasingly pronounced when only a few video samples are available. 
For example, videos of ``underwater diving" can exhibit significant intra-class variance from both static and dynamic perspectives. 
Static variability includes different types of diving equipment and diverse environments. 
Dynamic variability encompasses variations in motion speed, attitude adjustment, and technical style. 
The few-shot video data with significant individual differences are always far from enough to describe the real distribution of the action. 
In this case, the model may become overly specialized to the limited videos, resulting in overfitting and failing to generalize effectively to novel instances of the same action. 
In summary, the complexity of videos and the inevitable sample-wise variance within the same category make it challenging to learn novel action concepts with only a few labeled samples. 
With the rapid advancement of FSAR, a substantial body of new literature is being produced to address the unique challenges. 
Although recent relevant surveys~\cite{wang2020generalizing, li2023deep, gharoun2023meta, song2023comprehensive} have reviewed existing few-shot learning methods, 
they always overlook the unique characteristics of FSAR and cannot provide exhaustive summaries of recent research and findings in this field.
This motivates us to write a survey that reviews various few-shot action recognition methods from a comprehensive and systematic perspective, synthesizing the latest research developments as a guide for researchers and practitioners. 
Unlike early surveys that either briefly mention few-shot action recognition as a future trend or lack comprehensive content coverage, we deeply consider the unique challenges and provide an exhaustive analysis of all recent methods. 

Reviewing the development of few-shot action recognition, early approaches are \textbf{Generative-based} which focus on generating additional data to mitigate sample scarcity~\cite{mishra2018generative, kumar2019protogan, fu2019embodied, xian2021generalized}. 
In contrast, the majority of recent studies have embraced \textbf{meta-learning frameworks}, emphasizing critical stages including \textbf{Video Instance Representation}, \textbf{Category Prototype Learning}, and \textbf{Generalized Video Alignment}. 
Figure~\ref{fig:timeline} depicts a chronological overview of the development of different methods for few-shot action recognition.  
{\color{black}
As illustrated, there is a gradual rise in the quantity of FSAR publications, and generative-based methods are concentrated in the earlier period, whereas the subsequent techniques are primarily rooted in the meta-learning framework. 
It is also notable that methods based on multimodal pre-trained backbones emerged in 2023. 
}

\begin{figure}
    \centering
    \includegraphics[width=\linewidth]{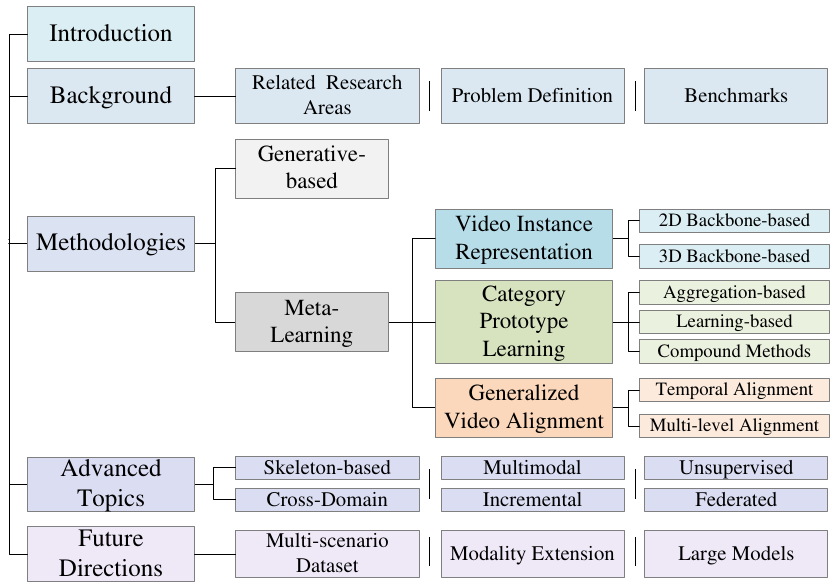}
    \caption{Overview of the organization for the survey. }
    \label{fig:organization}
\end{figure}

We systematically categorize existing works and develop a novel taxonomy (as shown in Figure~\ref{fig:organization}) to explore the contributions of existing methods, providing a thorough comparative analysis and discussion of recent advancements. 
Alongside this taxonomy, where existing FSAR approaches are categorized according to their contributions to addressing the challenges, we conduct a detailed analysis of the characteristics of existing approaches. 
Next, we highlight recent advancements in FSAR research within advanced fields, showcasing the methods developed to tackle application challenges. 
Finally, drawing from our comprehensive analysis, we propose promising research directions for future investigations. 
These insights are intended to inspire and guide future research endeavors, ultimately driving continued progress in the field. 

\textbf{Contribution. }
The main contributions of this survey can be summarized as follows: 
\begin{itemize}
  \item To the best of our knowledge, this is the first paper to provide a comprehensive review of the few-shot action recognition methods with a novel and systematic categorization, elucidating their achievements in the research field. 
  \item  Our investigation extends to advanced topics in few-shot action recognition, encompassing skeleton recognition, multimodal exploration, unsupervised learning, cross-domain learning, incremental learning, and federated learning. 
  By reviewing these advanced topics, our paper underscores the challenges and opportunities within these domains. 
  \item We discuss the principal research gaps in the current few-shot action recognition landscape and propose directions for future research, offering comprehensive insights and important guidance for further exploration within the academic community. 
\end{itemize}

\textbf{Organization. }
In this paper, we first review the related surveys in Section~\ref{sec:related survey}. 
Next, we examine the pertinent research areas, formulate the problem definition and the evaluation metrics, and introduce the benchmarks in Section~\ref{sec:background}. 
We then review the few-shot action recognition methods in Section~\ref{sec:methodologies}. 
Here, we organize previous methods into four types: generative-based methods, video instance representation learning, category prototype learning, and generalized video alignment. 
In Section~\ref{sec:advanced topics}, we discuss the latest developments in few-shot action recognition topics. 
Section~\ref{sec:future work} focuses on future research opportunities and discusses open issues. 
Finally, Section~\ref{sec:conclusion} concludes our paper. 
Our survey aims to contribute to the advancement of few-shot action recognition while providing valuable insights to the scholarly community.

\section{Related Surveys}
\label{sec:related survey}
\textbf{Action Recognition.} 
Kong \textit{et al.}\cite{kong2022human} and Saleem \textit{et al.}\cite{saleem2023toward} review contemporary deep learning methods for action recognition. 
The authors evaluate state-of-the-art techniques, providing a comparative analysis of technical difficulties and an examination of popular datasets. 
Additionally, Sun \textit{et al.}~\cite{sun2022human} review the current mainstream deep learning methods for both single and multiple data modalities, presenting comparative results on several multimodal benchmark datasets. 
However, these surveys provide only a brief introduction to the application of action recognition in few-shot scenarios. 
They lack a comprehensive discussion of the unique challenges and do not offer a thorough review of the advancements in this specialized field.

\textbf{Few-shot Action Recognition.} 
{\color{black}
Wang \textit{et al.}~\cite{wang2023survey} and Ruan \textit{et al.}~\cite{ruan2024advances} provide brief overviews of few-shot action recognition. 
Specifically, Wang \textit{et al.}~\cite{wang2023survey} summarized FSAR methods according to the task process, and Ruan \textit{et al.}~\cite{ruan2024advances} simply categorize FSAR methods into: memory networks, metric learning, and data augmentation. 
While these taxonomies bear a resemblance to general FSL surveys~\cite{li2023deep, gharoun2023meta}, they do not specifically highlight the contributions of current methodologies in addressing the unique challenges within FSAR. 
And these surveys fail to encompass all significant works since the establishment of this field and are limited in scope, lacking comprehensive comparative analysis, detailed taxonomy, and discussion of recent advancements.
On the contrary, we present the most systematic review in the field of FSAR, extending to advanced topics and broadening its applicability to more intricate contexts. It includes valuable details and fills the gaps left by existing surveys. 
}
\textbf{Zero-shot Action Recognition. } Estevam \textit{et al.}~\cite{estevam2021zero} present a survey of zero-shot action recognition methods that describes several techniques used to perform visual and semantic extraction. 
They do not take into account the few-shot action recognition, as zero-shot learning is inherently different from the few-shot setting. 
In zero-shot learning, predictions are made based on semantic information rather than on a few samples.

\textbf{Few-shot Learning.} 
Additionally, there are several surveys~\cite{wang2020generalizing, li2023deep, gharoun2023meta, song2023comprehensive} focus on few-shot learning. 
Wang \textit{et al.}\cite{wang2020generalizing} review the formal definition of few-shot learning and identify the core issue as the unreliability of the empirical risk minimizer. 
Gharoun \textit{et al.}~\cite{gharoun2023meta} and Song \textit{et al.}\cite{song2023comprehensive} further provide a novel taxonomy for recent advances in the field. 
Li \textit{et al.}~\cite{li2023deep} specialize in the few-shot classification task, offering a review of deep metric learning methods categorized according to three stages of metric learning.
These surveys briefly discuss the application of few-shot learning in action recognition as a future trend. 

\textbf{Meta-Learning.} 
As a prevalent solution for addressing data scarcity, meta-learning equips the learning system with the ability to acquire knowledge from multiple tasks, thereby enabling faster adaptation and generalization to new tasks. 
Hospedales \textit{et al.}\cite{hospedales2021meta} and Vettoruzzo \textit{et al.}\cite{vettoruzzo2024advances} describe the contemporary meta-learning landscape, illustrating how advancements in one area can benefit the field as a whole while preventing unnecessary duplication of efforts. 

Different from existing surveys, our study employs a comprehensive and systematic summary of various few-shot action recognition methods with an exhaustive discussion of their characteristics. 
Our research aims to form profound insights and identify significant future directions within the academic community.

\begin{figure}
    \centering
    \includegraphics[width=\linewidth]{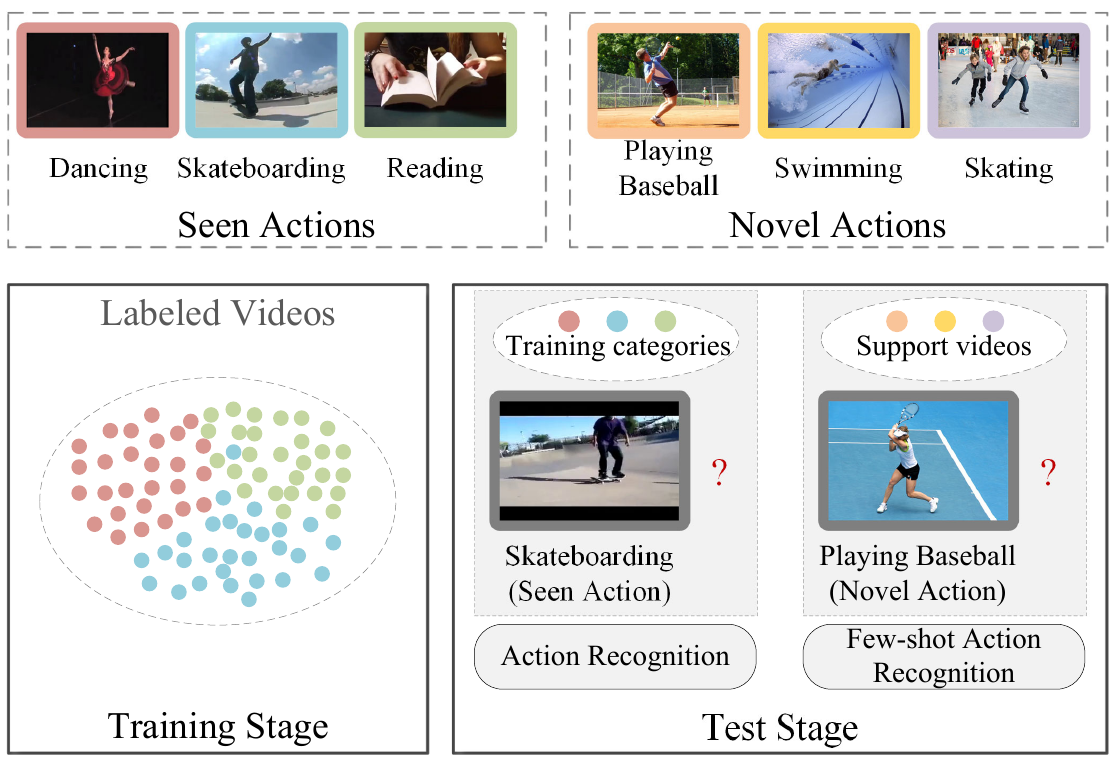}
    \caption{Comparison of the action recognition and few-shot action recognition tasks.}
    \label{fig:tasks}
\end{figure}

\section{Background}
\label{sec:background}
In this section, we provide a comprehensive research background for the few-shot action recognition. 
We introduce the relevant problems, provide a formal definition of the few-shot action recognition, and introduce the datasets leveraged in FSAR. 

\subsection{Relevant Research Areas}

Before defining the formulation of the few-shot action recognition task, it is essential to introduce related foundational problems, including human action recognition and few-shot learning. 
Understanding these prior problems will provide a better context for comprehending the intricacies of few-shot action recognition.

\subsubsection{Human action recognition}
Action recognition is a critical task in computer vision that involves identifying and categorizing human actions in video sequences. 
As illustrated in Figure~\ref{fig:tasks}, the task requires recognizing the categories of new videos by learning action concepts from a large number of labeled videos. 
Unlike static image classification, human action recognition must account for the temporal dynamics and sequential nature of actions, which significantly increases the complexity of the task. 
Common attempts leverage various advanced methods to accurately identify activities in video sequences, such as 3D Convolutional Neural Networks (CNNs)~\cite{ji20123d, taylor2010convolutional, tran2015learning}, 
RNN/LSTM networks~\cite{gupta2018social}, 
Attention mechanisms~\cite{li2020spatio}, 
Graph-based methods~\cite{si2019attention}, 
Transformers~\cite{liu2022video}, 
Self-supervised and unsupervised learning techniques~\cite{walker2014patch}. 
In recent years, the emergence of various sensors has enabled the representation of human actions through diverse modalities beyond RGB, including depth~\cite{roh2010view}, skeleton~\cite{wang2013learning, vemulapalli2014human}, infrared sequences~\cite{akula2018deep}, point clouds~\cite{min2020efficient}, event streams~\cite{innocenti2021temporal}, audio~\cite{liang2019audio}, acceleration~\cite{bayat2014study}, radar~\cite{lin2007doppler}, and WiFi~\cite{zou2019wifi}. 
Human action recognition has a wide range of applications and therefore has been attracting increasing attention in the field of computer vision, including surveillance~\cite{hu2007semantic}, human-computer interaction~\cite{barr2007video}, and sports analytics~\cite{soomro2015action}. 
Despite significant progress in efficient and accurate action recognition, further exploration of advanced few-shot action recognition is still required. 
Different from conventional supervised action recognition tasks, few-shot action recognition necessitates models to generalize to novel actions using only a limited number of videos.

\subsubsection{Few-shot Learning}
Few-shot action recognition is a critical application of few-shot learning due to the high cost of obtaining sufficient labeled videos. 
Specifically, Few-shot learning (FSL) is a technique that focuses on training models that can generalize effectively from a very limited number of labeled training examples. 
In contrast to traditional data-driven machine learning paradigms, FSL aims to enable models to perform well on new tasks with only a few instances per class by leveraging prior knowledge and inductive biases learned from related tasks or datasets. 
It is particularly valuable in scenarios where data collection and annotation are expensive or impractical. 
We mainly talk about the few-shot classification tasks such as image classification~\cite{vinyals2016matching, snell2017prototypical}, sentiment classification~\cite{yu2018diverse}, graph classification~\cite{ma2020adaptive}, and audio classification~\cite{wang2021audio}. 
% % 

Few-shot classification often employs an $N$-way $K$-shot framework, where $N$ represents the number of classes and $K$ denotes the number of examples per class. 
{\color{black}
In practical application, there are more universal settings:
Generalized FSL~\cite{xian2021generalized} encompasses the capability of models to recognize both few-shot categories and seen categories, reflecting real-world complexity. 
Open-set FSL~\cite{boudiaf2023open, liu2020few} aims to recognize few-shot classes and reject samples that do not belong to either seen or few-shot classes, with the goal of extending open-set recognition beyond the large-scale setting. 
Open-world FSL~\cite{shao2024collaborative, an2023instance} is a recent research field dedicated to accurately identifying target samples in scenarios where data is limited and labels are unreliable. 
This research holds significant practical implications and is highly relevant to real-world applications. 
The main challenge of these tasks is the same as the conventional few-shot setting: learning a new category with a few samples and identifying its category when given a new sample. 
Therefore, researching prevalent FSL algorithms is valuable and necessary. 
}

{\color{black}
Prominent FSL methodologies encompass meta-learning methods, data augmentation techniques, amortization-based methods, and in-context learning methods. }
Meta-learning is a widely used paradigm in FSL, which involves two phases: meta-training and meta-test. 
During meta-training, models are exposed to multiple episodes to learn a strategy for quick adaptation. 
During the meta-test, the model applies this strategy to new tasks with limited data. 
There are two types of meta-learning, including optimization-based methods~\cite{finn2017model, Liu2020ensemble}, which adjust the model parameters to facilitate fast learning on new tasks, and metric-based methods~\cite{vinyals2016matching, snell2017prototypical, Simon2020}, which focus on learning a distance metric to compare new examples with stored representations of known classes. 
Several other methods leverage data augmentation techniques to mitigate the issue of limited supervised information~\cite{zhang2018metagan, yang2021free}. 
These techniques involve generating synthetic data that augment the existing small dataset, thereby providing more examples for the model to learn from. 
{\color{black}
Additionally, amortization-based methods (e.g. CNPs~\cite{garnelo2018conditional}) combine the advantages of deep neural networks and Bayesian methods, enabling accurate predictions by observing only a handful of training data points.  
Recently, with the advancement of Large Language Models (LLMs), in-context learning (ICL) has been proposed to address FSL~\cite{hu2022context}. 
In this framework, an LLM receives a test instance and a few examples as input and directly performs recognition or reasoning without requiring any parameter updates. 
}
Compared to conventional few-shot classification tasks, few-shot action recognition is more challenging due to the additional temporal domain information in video data.

\begin{algorithm}[t]
\caption{Evaluation procedure of $N$-way $K$-shot action recognition}
\label{algorithm:test}
% \textbf{Require:} $\mathbb{D}_{test}$ \\
\KwIn{Novel dataset: $\mathbb{D}_{test}$; Learned Model: $\phi$;  Number of episodes: $E$. }
\KwOut{Mean Accuracy.}
\For{$\mathcal{T}^{(e)}$=$\mathcal{T}^{(1)}, \dots, \mathcal{T}^{(E)}$}{
  Randomly select $N$ novel classes. \\
  Randomly select $K$ samples from each class to construct the support set $\mathcal{S}^{(e)}=\{(x_i, y_i)\}_{_{i=1}}^{^{NK}}$. \\
  Randomly select $M$ samples from the remaining samples of $N$ classes as the query set $\mathcal{Q}^{(e)}=\{x_i\}_{_{i=1}}^{^{M}}$, with the labels $Y=\{y_i\}_{_{i=1}}^{^{M}}$. \\
  Record predicted labels $ \quad\quad \hat{Y}^{(e)} = \phi(\mathcal{Q}, \mathcal{S})$. \\
  Compute accuracy \\  $\quad\quad a^{(e)} = \frac{1}{M} \sum_{j=1}^{M}$ 
  $\mathds{1}$ $[\hat{y}_j^{(e)} = y_j^{(e)}]$.\\
  }
\Return mean accuracy $\frac{1}{E} \sum_{e} a^{(e)}$ 
\end{algorithm}
\subsection{Problem Definition}
\label{subsec:problem definition}

Few-shot action recognition refers to the task of effectively identifying human actions in unlabeled videos using a limited number of labeled samples from novel action categories (Figure~\ref{fig:tasks}). 
The task contains a base dataset $\mathbb{D}_{train}$, which includes ample labeled examples from base action classes $C_{train}$, and a novel dataset $\mathbb{D}_{test}$ from novel classes $C_{test}$ to assess the model's performance on previously unseen classes, ensuring $C_{test} \cap C_{train} = \varnothing$. 
In the evaluation phase, the $N$-way $K$-shot setting involves novel $N$ classes and $K$ examples from each class to form a support set, while an additional query set is used to evaluate the model's ability to generalize from the limited examples. 
As shown in Algorithm~\ref{algorithm:test}, the evaluation procedure includes many episodes. 
{\color{black} Each episode ($\mathcal{T}$)} is comprised of a query set $\mathcal{Q} \subset \mathbb{D}_{test}$ and a support set $\mathcal{S} \subset \mathbb{D}_{test}$. 
The query set $\mathcal{Q}=\{(x_i, y_i)\}_{_{i=1}}^{^{M}}$ contains $M$  query samples,
where $x_i$ denotes the video of the $i^{th}$ query sample $x_i$, and $y_i \in \{1, 2, ..., N\}$ denotes the class label.
The support set $\mathcal{S}=\{ (x_i, y_i) \}_{_{i=1}}^{^{NK}}$ contains ${K}$ samples for each of the $N$ classes.
Note that the class label of each query sample is invisible during the meta-test. 
The performance of the learning algorithm is measured by the classification accuracy averaged over all episodes. 

\subsection{Benchmarks}
\label{sec:benchmarks}
\begin{table*}[t]
\centering
\caption{Benchmarks for few-shot action recognition. }
\label{tab:dataset}
\resizebox{\linewidth}{!}{
\begin{tabular}{l|ccc|ccc|l|c|l}
\toprule 
\multirow{2}{*}{Benchmark} & \multicolumn{3}{c|}{Classes} & \multicolumn{3}{c|}{Videos} & \multirow{2}{*}{Reference} & \multirow{2}{*}{Perspective} & \multirow{2}{*}{Source} \\
 & Train  & Validation  & Test  & Train & Validation  & Test  &  &  & \\ \midrule\midrule
HMDB51 & 31 & 10 & 10  & 4280  & 1194 & 1292  & ARN ~\cite{zhang2020ARN} & Third Person & Movies, YouTube, Web  \\ 
UCF101 & 70 & 10 & 21  & 9154  & 1421 & 2745  & ARN~\cite{zhang2020ARN}  & Third Person & YouTube \\
Kinetics & 64 & 12 & 24  & 6400  & 1200 & 2400  & CMN~\cite{zhu2018CMN} & Third Person & YouTube \\
SSv-full & 64 & 12 & 24  & 77500 & 1925 & 2854  & OTAM~\cite{cao2020OTAM}  & First Person & Crowd-Sourced \\ 
SSv2-small & 64 & 12 & 24  & 6400  & 1200 & 2400  & LIM~\cite{zhang2021LIM}  & First Person & Crowd-Sourced \\ 
EPIC-Kitchens  & 60 & -  & 20  & 66056 & -  & 10646 & HyRSM~\cite{wang2022hybrid}  & First Person & Crowd-Sourced  \\ \bottomrule 
\end{tabular}}
\end{table*}
In this section, we briefly introduce benchmark datasets for few-shot action recognition. 
Statistics of the datasets and commonly used experimental settings are listed in Table~\ref{tab:dataset}. 
% , and sample videos are shown in Fig.~\ref{}. 
% 
% 
% UCF101

\textbf{UCF101. }
The UCF101 dataset~\cite{soomro2012ucf101} is a widely recognized benchmark in the field of human action recognition.
These categories span five broad types: Human-Object Interaction, Body-Motion Only, Human-Human Interaction, Playing Musical Instruments, and Sports. 
For few-shot action recognition, the widely used split benchmark is proposed by ARN~\cite{zhang2020ARN}, where the training, validation and test sets consist of 70, 10 and 21 classes and 9154, 1421 and 2745 videos respectively. 
% 

% HMDB51
\textbf{HMDB51. }
The HMDB51 dataset~\cite{kuehne2011hmdb} is a prominent benchmark for human action recognition in video sequences, containing 6,766 video clips categorized into 51 distinct action classes. 
Each video clip in HMDB51 is collected from a diverse set of sources, including movies, public databases, and YouTube, ensuring a wide variety of scenarios, backgrounds, and camera motions. 
The few-shot benchmark established by ARN is structured with a division of 64 action classes and 4280 videos for training, 12 classes and 1194 videos for validation, and 24 classes and 1292 videos for testing.

% Kinetics
\textbf{Kinetics.}
\begin{figure}[t]
    \centering
    \includegraphics[width=0.85\linewidth]{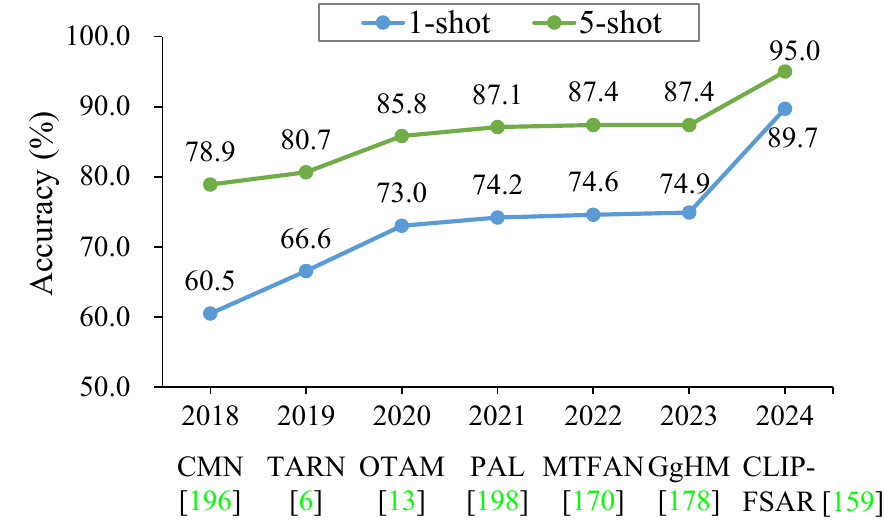}
    \caption{Performance of typical few-shot action recognition methods on Kinetics dataset~\cite{carreira2017kinetics} in 5-way 1-shot and 5-way 5-shot settings.}
    \label{fig:performance}
\end{figure}
The Kinetics dataset~\cite{carreira2017kinetics} is one of the most comprehensive and influential benchmarks for human action recognition, consisting of 306245 high-quality video clips. 
The dataset has been released in several versions, including Kinetics-400~\cite{kay2017kinetics-400}, Kinetics-600~\cite{carreira2018kinetics-600}, and Kinetics-700~\cite{carreira2019kinetics-700}, the new version increases the number of classes and video samples to enhance the dataset's comprehensiveness. 
The few-shot benchmark proposed by CMN~\cite{zhu2018CMN} includes 100 randomly selected classes from the Kinetics dataset, with each class containing 100 examples. 
These 100 classes are divided into 64/12/24 non-overlapping classes with 6400/1200/2400 videos designated for the training set, validation set, and test set respectively. 
Figure~\ref{fig:performance} displays the performance trends of state-of-the-art methods over the years.

% Something to Something v2
\textbf{Something-Something V2.}
The Something-Something V2 dataset~\cite{goyal2017something} is a significant benchmark in the field of video understanding and human action recognition. 
It consists of over 220,000 video clips covering 174 action classes, making it one of the largest and most diverse datasets for action recognition. 
Currently, there exist two widely used few-shot benchmarks, both containing 100 action categories with 64/24/12 classes for meta-training, meta-validation, and meta-test, though differing in action categories and sample sizes. 
SSv2-full, proposed by OTAM~\cite{cao2020OTAM}, includes all samples from the original dataset for each class. 
SSv2-small, proposed by LIM~\cite{zhang2021LIM}, comprises 100 samples per class.

% EPIC-Kitchens
\textbf{EPIC-Kitchens.}
The EPIC-Kitchens dataset~\cite{damen2018EPIC-Kitchens} is a large-scale benchmark specifically designed for fine-grained action recognition in egocentric (first-person) video recordings. 
It comprises over 55 hours of unscripted activities captured in kitchen environments by 32 participants from diverse cultural backgrounds, resulting in nearly 40,000 action segments. 
These segments are annotated with detailed labels, covering 125 verb classes and 352 noun classes, which describe various cooking-related actions and objects.
HyRSM~\cite{wang2022hybrid} collected the few-shot benchmark where the training and test sets contain 60/20 action classes and 66056/10646 videos respectively.

\section{Methodologies}
\label{sec:methodologies}
\begin{figure}[t]
    \centering
    \includegraphics[width=\linewidth]{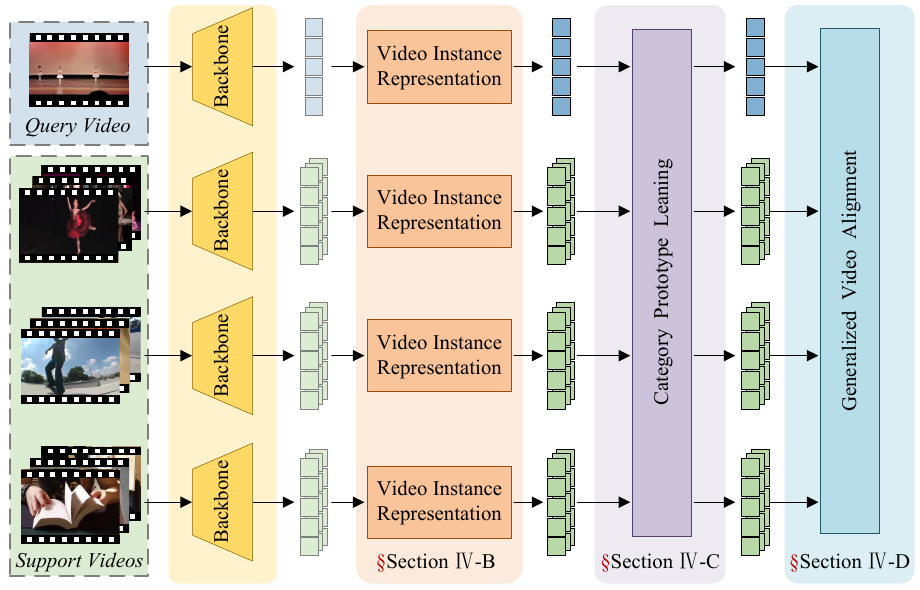}
    \caption{
    The general framework of few-shot action recognition with meta-learning.
    % , which includes the episode preparation, feature embedding extraction, video feature enhancement, prototype learning and video alignment. 
    }
    \label{fig:framework}
\end{figure}

In order to mitigate the challenge of overfitting or non-convergence made by insufficient samples, early methods propose to estimate more accurate novel class distributions by generating more sample features based on a few available samples. 
Reviewing the development of few-shot action recognition, early approaches are generative-based which focus on generating additional data to mitigate sample scarcity~\cite{mishra2018generative, kumar2019protogan, fu2019embodied, xian2021generalized}. 
Whereas, the majority of solutions proposed for FSAR belong to the meta-learning framework. 
{\color{black}
These methods adopt ProtoNet as their backbone framework, a metric-based meta-learning approach that simulates few-shot scenarios during training to enable rapid generalization from a few samples, rather than directly memorizing action features. In contrast to other meta-learning methods (e.g., MAML~\cite{finn2017model}), ProtoNet-based methods have lower computational costs and are easier to optimize. 
Furthermore, they facilitate the expansion of distance metrics to enhance spatiotemporal matching performance for FSAR. }

{\color{black}
Figure~\ref{fig:framework} offers an overview of the meta-learning framework for few-shot action recognition, involving five main stages. }
To begin, a meta-training/meta-test episode is created containing  $N$-way $K$-shot support videos of the novel category, along with query videos. 
Videos are then processed to extract frame-level or video-level features. 
Subsequently, through the enhancement of video features, a more robust and generalized video representation is achieved. 
Next, the prototype representations for various categories are computed. 
Finally, an alignment strategy is then designed to measure the similarity between the query and prototype, leading to the classification results. 
In this framework, three critical issues need to be primarily considered:
(1) Learning video instance representations with transferable ability to enhance the quality and discriminative power of feature embeddings, thereby capturing complex, data structures in few-shot scenarios. 
(2) Creating accurate action prototypes from a limited number of videos to facilitate reliable comparison and classification.
(3) Optimizing distance metrics to effectively measure similarity between support and query videos in a manner that benefits few-shot learning tasks, ensuring that the learned metrics generalize well to novel classes. 
In the remainder of this section, we provide a review of existing approaches, categorized into four groups based on the aspects they focus on.

\subsection{Generative-based methods}
We first review the early methods that adopt the generative-based learning paradigm. 
These methods use general deep learning strategies to solve the few-shot action recognition and to mitigate the challenge of overfitting or non-convergence made by insufficient samples. 
{\color{black}
Several methods~\cite{yang2021free} leverage prior knowledge of the distribution of seen classes to create pseudo samples for novel classes. 
Although these pseudo-samples may not fully encapsulate the class distribution, they can aid in mitigating the distribution bias caused by limited novel samples to several extent. 
}
However, due to the complexity of video features, generating video features faces greater challenges. 
While image generation has made significant progress, generating videos for training deep learning architectures remains challenging due to the high dimensionality of the output space. 

Mishra \textit{et al.}~\cite{mishra2018generative} propose a method for achieving a seamless transition from zero-shot to few-shot action recognition. 
They estimate the initial Gaussian distributions for unseen classes using a zero-shot method. 
When labeled data for these unseen classes becomes available, the model updates the parameters of these distributions, refining the estimates and enabling the synthesis of novel examples by sampling from the respective class distribution. 
ProtoGAN~\cite{kumar2019protogan} synthesizes additional examples for novel categories by conditioning a generative adversarial network with class prototype vectors. 
These vectors, learned via a class prototype transfer network from examples of seen categories, generate examples semantically akin to real instances of the novel class. 
Xian \textit{et al.}~\cite{xian2020generalized} utilize user-tagged videos in YFCC100M~\cite{thomee2016yfcc100m} dataset to synthesize variations of the training data conditioned on novel classes.  
They advocate for using generative models to sample low-dimensional video features from learned distributions, leveraging semantic embeddings of class names like Word2Vec~\cite{mikolov2013efficient} or BERT~\cite{devlin2018bert} to control the generation process. 
This approach facilitates knowledge transfer between base and novel classes by capturing class similarities from the language modality. 
In summary, while these generative-based methods offer benefits in avoiding overfitting and enhancing robust representations, they come with additional computational overhead and cannot provide competitive performance.

%%%%%%%%%%%%%%%%%%%%%%%%%%%%%%%%%%%%%%%%%%%%%%%%%%%%%%%%%%%%%%%%% 
\subsection{Video Instance Representation}
% 
% 
% Learning video instance representations necessitates networks adept at extracting discriminative spatiotemporal video features and exhibiting a strong generalization ability to novel actions. 

Few-shot action recognition requires learning video instance representations that are spatiotemporally discriminative and can exhibit strong generalization ability to novel actions. 
Since few-shot learning in the image domain has made significant progress, the spatial modeling techniques employed in these methods can potentially enhance the features of video data. 
Nonetheless, identifying novel action concepts from complex video data presents unique challenges attributable to the inherent temporal dimension in videos.
Further investigation into modeling video features with enhanced generalization capabilities is imperative. 
Numerous works in FSAR literature have introduced innovative approaches {\color{black}to extract} robust features from diverse perspectives. 
Previous methods mainly rely on two kinds of backbones for extracting representation embeddings from videos: 2D backbones and 3D backbones. 

\subsubsection{2D Backbone-based}

% \subsubsection{2D models}
Many methods~\cite{zhu2018CMN, cao2020OTAM, bo2020few, perrett2021temporal, li2022ta2n, thatipelli2022spatio} are inspired by well-established supervised action recognition techniques~\cite{Simonyan2014two-stream, wang2016temporal} to extract frame-level features using 2D convolutional networks, which are fundamental operations in the image domain. 
These approaches represent videos as sequences of feature vectors corresponding to each frame. 

Commonly used networks include ResNet~\cite{he2016resnet}, GoogLeNet~\cite{szegedy2015googlenet}, and ViT~\cite{dosovitskiy2020vit}, all pre-trained on large-scale image datasets such as ImageNet~\cite{Deng2009Imagenet}.
For instance, AFO~\cite{zou2020adaptation} addresses video feature extraction by pre-training GoogLeNet~\cite{szegedy2015googlenet} on seen classes and projecting the output into an adaptation-oriented feature space by fusing important feature dimensions. 
Recently, with the widespread adoption of large-scale models, researchers~\cite{wang2023clip, ni2024multimodal, tang2024semantic} have explored the application of the visual large model~\cite{tong2022videomae} or language-image pretrained model~\cite{radford2021CLIP} in FASR. 
Leveraging the rich representations and strong generalization capabilities of these models, these methods have demonstrated significant performance improvements. 
In few-shot action recognition, relying solely on frame-level features extracted by 2D networks is insufficient for reliable action recognition, as it does not account for the complex temporal information in videos and the need for generalization ability. 
Recently, Huang \textit{et al.}~\cite{huang2024manta} introduce 
Mamba architecture into FSAR for enhancing local features from fragments of a long sub-sequence and performing implicit temporal alignment. 
Therefore, many methods propose new techniques to learn video instance representations by enhancing frame-level features. 
\textbf{CNN Models. }
% GoogleNet
% ResNet50
% ResNet50
Several early works~\cite{xu2018dense, bo2020few, wang2021temporal} propose to employ Convolutional Neural Networks (CNNs) to model long-term relations in videos by extracting temporal dependencies. 
%  % 
However, these CNN-based methods face limitations such as inflexibility in modeling variable-length videos and inadequate representation of dynamic features in FSAR. 
% 
% As a result, their application was confined to early approaches in the field of few-shot action recognition. 

\textbf{GNN Models. }
As an effective solution for learning and inference on complex relational data, Graph Neural Network (GNN) models are designed to process data structured as graphs. 
Several studies~\cite{hu2021learning, feng2022learning, xing2023boosting, wang2023task} have adopted GNNs to address temporal modeling in few-shot action recognition. 
% 
% ResNet50
For example, Hu \textit{et al.}~\cite{hu2021learning} propose a dual-pooling graph neural network with hierarchical customized graph pooling layers to enhance the discriminative ability in accurately selecting representative video content. 
The authors construct intra-video graphs to identify the most representative frames and an inter-video graph to adaptively eliminate negative relations between videos, thereby improving overall video representations. 
% 
% 
% ResNet50
Similarly, Feng \textit{et al.}~\cite{feng2022learning} utilize GNNs to analyze local-global relations in videos, generating hierarchical capsules for understanding video relationships and structural information. 
The approach incorporates inter-video and intra-video routing processes to filter low-discriminative capsules comprehensively, ensuring efficient representations. 
% 
% ResNet50
Additionally, Wang \textit{et al.}~\cite{wang2023task} leverage GNNs to explore intra-class and inter-class relations across video pairs simultaneously, enhancing task-aware video representations.
In summary, the capability of GNNs to model relational structures facilitates exploring associations between frames or samples, thereby enhancing the robustness of feature representation.

\textbf{Attention Strategies. }
The attention mechanism~\cite{vaswani2017attention} is another effective solution for modeling temporal information, as it dynamically focuses on the most relevant parts of a sequence, thereby capturing long-range dependencies and enhancing the representation of temporal dynamics. 
Many FSAR methods apply attention strategies to enhance the representation of video instances. 
% 
% ResNet
TAVs~\cite{bo2020few} adapt videos of varying lengths to preserve their temporal information and encode temporal relationships across the entire video by assigning importance scores to training parameters, facilitating a more accurate capture of temporal patterns compared to deep networks trained with limited data.
% 
% ResNet
As well, Wang \textit{et al.}~\cite{wang2021few} introduce bidirectional attention which imitates the top-down and bottom-up attention mechanism to capture subtle action details from extracted features. 
The top-down attention is directly driven by few-shot action recognition tasks, capturing specific action-related features like sports equipment and critical body parts, while the bottom-up attention is supervised by predicted saliency volumes, focusing on class-agnostic saliency features like athletes and objects in motion. 
% 
% 
% 
% ResNet
Inspired by the idea of neural architecture search, Cao \textit{et al.}~\cite{cao2022searching} propose to explore optimal architectures concerning the sequence order of temporal attention and spatial attention blocks for few-shot action recognition. 
The authors employ a pre-defined transformer space to incorporate both temporal and spatial attention while implementing a space-shrinking strategy along with the analysis of training losses to examine the emphasis on spatial and temporal information at different stages, aiming for superior architectures with enhanced accuracy and cost efficiency. 
% 
% 
% 
% 
% ResNet
Additionally, SloshNet~\cite{xing2023revisiting} utilizes the attention mechanism to encode complementary global and local temporal representations. 
% 
% ResNet
Zhang \textit{et al.}~\cite{zhang2023importance} observe that spatial misalignment between objects is more prevalent in videos compared to temporal inconsistencies. 
Recognizing the significance of spatial relations, they propose an improved few-shot action recognition approach that integrates both spatial and temporal information. 
Specifically, they introduce a novel spatial alignment cross Transformer, which not only learns to realign spatial relations but also incorporates temporal information. 
Kumar \textit{et al.}~\cite{kumar2024trajectory} leverage space-time attention mechanism to assimilate the point trajectory extracted from videos, and model inter-point relationships for consolidating spatio-temporal information. 

These methods leverage the sequence-dependent modeling capability of the attention mechanism to focus on key spatiotemporal features in videos, thereby improving the robustness of video instance representations. 

\textbf{Auxiliary Information Enhancement Techniques. }
Several methods leverage augmentation strategies to enhance the robustness and discriminative power of video representations at the feature level, utilizing auxiliary information such as self-supervised signals, motion information, semantic information, and additional unlabeled videos. 
The auxiliary information serves as guidance or supplementation for video representations.

In action recognition, depth, and motion information have been proven to be effective guidance signals, helping enhance contextual understanding, capturing temporal dynamics, reduces bias~\cite{wang2012mining, fathi2008action}. 
This idea has been leveraged in few-shot action recognition as well~\cite{fu2020depth, wu2022motion, luo2022long, wang2023molo}. 
% 
% guide (depth, semantic)
% resnet
For example, AMeFu-Net~\cite{fu2020depth} discovers the scene representation bias, which is inevitably learned by supervised video recognition methods since human actions often happen in specific scene contexts. 
Therefore, the authors introduce depth information as a carrier of the scene for extra visual information. 
% 
% 
% 
% motion 
% resnet
Wu \textit{et al.}~\cite{wu2022motion} proposes to extract the task-specific motion patterns of videos from the temporal differences across consecutive frames. 
This approach aims to discover and enhance task-shared informative motion information, thereby facilitating improved alignment between videos within the same task. 
% 
% 
% motion
% resnet
Luo \textit{et al.}~\cite{luo2022long} leverage compressed domain knowledge~\cite{calderbank2009compressed} and propose a Long-Short Term Cross-Transformer for few-shot video classification where the motion vectors encode temporal cues and indicate the importance of each frame. 
% 
% 
% motion
% resnet
Wang \textit{et al.}~\cite{wang2023molo} develop a motion-augmented long-short contrastive learning method to jointly model the global contextual information and motion dynamics. 
To explicitly integrate the global context into the local matching process,  a long-short contrastive objective is applied to enforce frame features to predict the global context of the videos that belong to the same class. 
The authors take frame differences as motion information and design a motion autoencoder to explicitly extract motion features between frame representations by reconstructing pixel motions. 

Beyond visual information, semantic information is also commonly used to enhance video representations and improve generalization to novel actions. 
Several methods focus on leveraging the semantic knowledge of action concepts~\cite{shi2022knowledge, xia2023few, kim2024leveraging}.
% 
% % semantic
% resnet
For instance, Xia \textit{et al.}~\cite{xia2023few} tackles the deficiency of semantic information by introducing external knowledge learned globally to enhance video representations learned locally within individual tasks. 
Specifically, the authors maintain a global meta-action bank that stores representations of atomic action snippets, which can be used to compose various complex action categories. 
This global meta-action bank is shared across all action types and individual tasks, enriching the video feature representations learned locally in each task through a novel single-value decomposition technique.
% 
% semantic
% resnet
Similarly, Shi \textit{et al.}~\cite{shi2024commonsense} leverage commonsense knowledge of actions from external resources to prompt-tune a powerful pre-trained vision-language model for few-shot classification. 
A large-scale corpus of language descriptions of actions are collected as text proposals, to build an action knowledge base. 
The collection of text proposals is done by filling in a handcraft sentence template with an external action-related corpus or by extracting action-related phrases from captions of Web instruction videos. 
CLIP~\cite{radford2021CLIP} is leveraged to generate matching scores of the proposals for each frame, and the scores can be treated as action semantics with strong generalization. 
% 

% unlabeled videos
% resnet
% 

The scarcity of labeled videos poses a bottleneck for FSAR and leveraging readily available unlabeled videos for distribution estimation and representation learning is a viable solution. 
LIM~\cite{zhu2020label} proposes a semi-supervised approach where millions of unlabeled videos are utilized during training. 
This method incorporates a label-independent memory module to store label-related features, enabling the generation of robust class prototypes essential for few-shot training.

In summary, auxiliary information enhancement approaches offer substantial benefits by enriching video representations with diverse data modalities and semantic guidance. 
However, these methods often require additional computational resources to achieve enhanced performance.

{\color{black}
\textbf{Finetuning Strategies. }
The majority of the previously mentioned methods utilize models pre-trained on image dataset as a backbone, with fine-tuning all parameters in the backbone~\cite{li2024learning}. 
In recent years, there have been emerging multimodal pre-trained models typified by the CLIP~\cite{radford2021CLIP}, which generally do not require extensive fine-tuning. 
Recently, several FSAR methods~\cite{wang2023clip} adopt the CLIP model as their backbone and proposed specific fine-tuning approaches tailored to CLIP to enhance the video feature extraction capabilities for the FSAR task. 
% 
% Multimodal Prototype-Enhanced Network for Few-Shot Action Recognition
% CLIP-guided Prototype Modulating for Few-shot Action Recognition
% Multi-view Distillation based on Multi-modal Fusion for Few-shot Action Recognition(CLIP-M2DF)
For instance, several studies~\cite{wang2023clip, ni2024multimodal, guo2024multi, wu2024efficient} froze the parameters of the text encoder to preserve its generalization capabilities, while fine-tuning the parameters of the visual encoder to adapt it to the video domain.
% 
% Seeing in Flowing: Adapting CLIP for Action Recognition with Motion Prompts Learning
Additionally, Wang \textit{et al.}~\cite{wang2023seeing} expand the original encoders of CLIP with light-weight extensions to transfer the pre-trained model into action recognition task while maintaining the generalization of CLIP via freezing the parameters of the image encoder. 
Several studies propose to adopt the adapter into the encoder in CLIP to adapt the model to the FSAR data. 
% 
% MA-FSAR: Multimodal Adaptation of CLIP for Few-Shot Action Recognition
% 
{\color{black}
For example, Xing \textit{et al.}~\cite{xing2023multimodal} propose the Fine-grained Multimodal Adaptation for fine-tuning the visual encoder in CLIP. 
This strategy is designed to capture global motion cues enhance local temporal dynamics and highlight fine-grained semantics related to actions.}
Differently, Li \textit{et al.}~\cite{li2024frame} propose the perceiver-based adapter to recurrently capture the sequential dynamics alongside the timeline, which could perceive the order change.  
% 
% Task-Adapter: Task-specific Adaptation of Image Models for Few-shot Action Recognition
% 
{\color{black}
Cao \textit{et al.}~\cite{cao2024task} introduce the Task-Adapter into the final layers of the visual encoder while maintaining the frozen parameters of the original pre-trained model. }
Within each Task-Adapter, the frozen self-attention layer is repurposed to carry out task-specific self-attention across various videos within the assigned task. This process enables the capture of both unique information among classes and shared information within classes, thus facilitating task-specific adaptation and enhancing subsequent metric measurements between the query feature and support prototypes.
% 
% D2ST-Adapter: Disentangled-and-Deformable Spatio-Temporal Adapter for Few-shot Action Recognition
% 
D$^2$ST-Adapter~\cite{pei2023d} is formulated with a dual-pathway architecture to distinctly encode spatial and temporal features. 
This design allows for customization with anisotropic sampling densities along spatial and temporal domains, facilitating the dedicated learning of spatial and temporal features within respective pathways.

}

\subsubsection{3D Backbone-based}
% 
% 
% \subsubsection{3D models}
Several methods~\cite{kumar2019protogan, bishay2019tarn,zhang2020ARN, bo2020few, patravali2021unsupervised} draw inspiration from widely used video classification techniques~\cite{taylor2010convolutional, ji20123d, tran2015learning, carreira2017kinetics, hara2018resnet3d}, utilizing pre-trained 3D models as the backbone to capture short-term dependencies within videos, such as C3D~\cite{ji20123d}, I3D~\cite{carreira2017kinetics}, R3D~\cite{tran2018R3D} and VideoMAE~\cite{tong2022videomae}. 
However, relying solely on 3D models may overlook long-term temporal dependencies in videos. 
Moreover, the challenge persists in generalizing 3D models, which may exhibit biases towards specific patterns, to a limited number of novel class video samples.

To address these challenges, TARN~\cite{bishay2019tarn} proposes leveraging a bidirectional Gated Recurrent Unit (GRU) layer that utilizes local features extracted by the pre-trained C3D network~\cite{ji20123d} to learn globally-aware features, enabling each time step to access both backward and forward information across the entire video.
Similarly, ARN~\cite{zhang2020ARN} employs a pre-trained C3D network~\cite{ji20123d} to effectively capture short-range action patterns, which are then aggregated using permutation-invariant pooling into fixed-length representations. 
Additionally, the approach employs self-supervision to apply augmentation patterns for robust video representation, enhancing the training of a more resilient encoder.

% 
% semantic
% 
% 
Recently, Tang \textit{et al.}~\cite{tang2024semantic} extract global-wise video features using VideoMAE~\cite{tong2022videomae}, which captures intricate spatio-temporal relationships within videos. 
Furthermore, they encoded textual semantics into the video representation, adaptively fusing features from text and video to encourage the visual encoder to extract more semantically consistent features.

\subsubsection{Discussion}

The performances of different video instance representation methods are illustrated in Table~\ref{tab:video}. 
We observe that contemporaneous 3D backbone-based methods often demonstrate superior performance compared to those based on 2D backbones. 
For instance, Liu \textit{et al.}~\cite{liu2023mastaf} evaluated their proposed method using both 2D and 3D backbones, observing that the performance with a 3D CNN embedding network surpassed that of a 2D CNN embedding network.
This advantage arises because 3D backbones are primarily pretrained on video datasets, enabling them to effectively capture temporal dependencies between frames. 
However, 3D backbones entail high computational complexity, limiting their practical applicability in real-world scenarios.
In contrast, 2D backbone-based methods are preferred for their flexibility and efficiency in handling spatial information, accommodating variable-length videos, and maintaining low computational complexity.

\begin{table*}[t]
\caption{Results of different video instance representation learning methods.}
\label{tab:video}
\setlength{\extrarowheight}{0pt}
\setlength{\tabcolsep}{2pt}
\resizebox{\linewidth}{!}{
\begin{tabular}{llllcccccccccc}
\toprule
\multirow{2}{*}{Category}  &\multirow{2}{*}{Method}  & \multirow{2}{*}{Publication} & \multirow{2}{*}{Backbone} & \multicolumn{2}{c}{SSv2-full} & \multicolumn{2}{c}{Kinetics} & \multicolumn{2}{c}{UCF} & \multicolumn{2}{c}{SSv2-small} & \multicolumn{2}{c}{HMDB} \\
&  &  &  & 1-shot  & 5-shot  & 1-shot  & 5-shot  & 1-shot  & 5-shot  & 1-shot  & 5-shot  & 1-shot  & 5-shot  \\ \midrule
\multirow{22}{*}{\begin{tabular}[l]{@{}l@{}}2D   \\      backbone-\\      based\end{tabular}} & TAVs~\cite{bo2020few}  & WACV'2020  & ResNet-152  & -  & -  & -  & -  & -  & 83.2  & -  & -  & -  & 58.9  \\
& CMN-J~\cite{zhu2020label}  & TPAMI'2020  & ResNet-50  & -  & -  & 60.5  & 78.9  & -  & -  & 36.2  & 48.8  & -  & -  \\
& AOF~\cite{zou2020adaptation}  & TMM'2020  & GoogLeNet  & -  & -  & -  & -  & 64.1  & -  & -  & -  & 56.3  & -  \\
& AMeFu-Net~\cite{fu2020depth}  & ACM MM'2020  & ResNet-50  & -  & -  & 74.1  & 86.8  & 85.1  & 95.5  & -  & -  & 60.2  & 75.5  \\
& DPGNN~\cite{hu2021learning}  & TMM'2021  & ResNet-50  & -  & -  & 66.8  & 80.7  & 67.7  & 84.7  & -  & -  & -  & -  \\
& Wang \textit{et al.}~\cite{wang2021semantic}  & ACM MM'2021  & ResNet-50  & -  & -  & 75.2  & 87.1  & 86.5  & 95.8  & -  & -  & 61.6  & 76.2  \\
& Liu \textit{et al.}  & ACM MM'2022  & ResNet-50  & 47.1  & 61.6  & 73.6  & 86.2  & 83.5  & 96.0  & -  & -  & 59.9  & 73.5  \\
& SST~\cite{cao2022searching}  & NeurIPS'2022  & ResNet-50  & -  & -  & -  & -  & 65.4  & 70.4  & -  & -  & 52.4  & 62.2  \\
& Luo \textit{et al.}~\cite{luo2022long}  & IJCAI'2022  & ResNet-50  & 46.7  & 66.7  & 73.4  & 86.5  & 85.7  & 96.5  & -  & -  & 60.9  & 76.8  \\
& STRM~\cite{thatipelli2022spatio}  & CVPR'2022  & ResNet-50  & 43.1  & 68.1  & 62.9  & 86.7  & 80.5  & 96.9  & 37.1  & 55.3  & 52.3  & 77.3  \\
& DR-CapsGNN~\cite{feng2022learning} & TMM'2023  & ResNet-50  & 41.2  & 51.7  & 72.5  & 83.4  & -  & -  & -  & -  & -  & -  \\
& SloshNet~\cite{xing2023revisiting} & AAAI'2023  & ResNet-50  & 46.5  & 68.3  & -  & 87.0  & -  & 97.1  & -  & -  & -  & 77.5  \\
& SA-CT~\cite{zhang2023importance}  & ACM MM'2023  & ResNet-50  & 48.9  & 69.1  & 71.9  & 87.1  & 85.4  & 96.4  & -  & -  & 60.4  & 78.3  \\
& RFPL~\cite{xia2023few}  & ICCV'2023  & ResNet-50  & 47.0  & 61.0  & 74.6  & 86.8  & 84.3  & 92.1  & -  & -  & -  & -  \\
& TADRNet~\cite{wang2023task}  & TCSVT'2023  & ResNet-50  & 43.0  & 61.1  & 75.6  & 87.4  & 86.7  & 96.4  & -  & -  & 64.3  & 78.2  \\
& MoLo~\cite{wang2023molo}  & CVPR'2023  & ResNet-50  & 56.6  & 70.6  & 74.0  & 85.6  & 86.0  & 95.5  & 42.7  & 56.4  & 60.8  & 77.4  \\
& MRLN~\cite{wang2024few}  & TCSVC'2024  & ResNet-50  & 45.9  & 61.3  & 75.7  & 87.6  & 86.9  & 96.3  & -  & -  & 65.5  & 78.6  \\
& Kumar \textit{et al.}~\cite{kumar2024trajectory} & ECCV'2024  & ViT  & 57.7  & 74.6  & 81.9  & 91.1  & 92.0  & 95.5  & 47.9  & 64.4  & 60.0  & 77.0  \\
& Shi \textit{et al.}~\cite{shi2024commonsense}  & TMM'2024  & CLIP-ViT-B  & 44.7  & 62.4  & 85.2  & 94.3  & 97.4  & 99.4  & -  & -  & 75.8  & 87.4  \\
& TSAM~\cite{li2024frame}  & ArXiv'2024  & CLIP-ViT-B  & 65.8  & 74.6  & 96.2  & 97.1  & 98.3  & 99.3  & 60.5  & 66.7  & 84.5  & 88.9  \\
& MA-FSAR~\cite{xing2023multimodal}  & ArXiv'2024  & CLIP-ViT-B  & 63.3  & 72.3  & 95.7  & 96.0  & 97.2  & 99.2  & 59.1  & 64.5  & 95.7  & 96.0  \\
& Task-Adapter~\cite{cao2024task}  & ACM MM'2024  & CLIP-ViT-B  & 71.3  & 74.2  & 95.0  & 96.8  & 98.0  & 99.4  & 60.2  & 70.2  & 83.6  & 88.8  \\
& Manta~\cite{huang2024manta}  & ACM MM'2024  & ImageNet-VM  & 66.1  & 89.1  & 84.4  & 96.2  & 96.9  & 99.4  & -  & -  & 89.1  & 96.6  \\ \midrule
\multirow{4}{*}{\begin{tabular}[l]{@{}l@{}}3D   \\      backbone-\\      based\end{tabular}} & ARN~\cite{zhang2020ARN}& ECCV'2020  & C3D  & -  & -  & 63.7  & 82.4  & 66.3  & 83.1  & -  & -  & 45.5  & 60.6  \\
& Zhang \textit{et al.}~\cite{zhang2022multi}  & TMM'2022  & C3D  & -  & -  & 66.3  & 85.2  & 68.2  & 87.1  & -  & -  & 46.7  & 60.3  \\
& MASTAF~\cite{liu2023mastaf}  & WACV'2023  & ViViT  & 60.7  & -  & -  & -  & 91.6  & -  & 45.6  & -  & 69.5  & -  \\
& SAFSAR~\cite{tang2024semantic}  & WACV'2024  & VideoMAE  & 74.7  & 78.7  & -  & -  & 99.2  & 99.6  & 60.7  & 64.9  & 78.4  & 85.7  \\ \bottomrule
\end{tabular}
}
\end{table*}

%%%%%%%%%%%%%%%%%%%%%%%%%%%%%%%%%%%%%%%%%%%%%%%%%%%%%%%%%%%%%%%%%
\subsection{Category Prototype Learning}
Another challenge of few-shot action recognition is learning prototypes for novel action concepts from a few videos.
It involves learning an embedding space where video samples cluster around a single prototype for each action class. 
Action recognition is then performed by identifying the nearest class prototype for an embedded query video. 

\begin{figure}
    \centering
    \includegraphics[width=\linewidth]{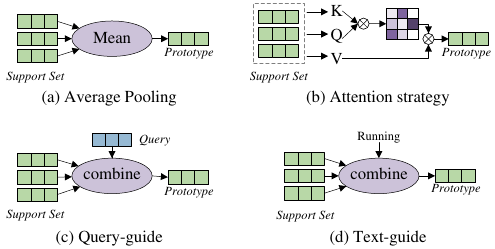}
    \caption{
    Typical prototype aggregation schemes: 
    (a) Average Pooling~\cite{bishay2019tarn, zhang2020ARN};
    (b) Attention Strategy~\cite{zhu2021few, wang2021temporal};
    (c) Query-guided Combination~\cite{perrett2021temporal};
    (d) Text-guided Combination~\cite{ni2024multimodal, wang2023clip}. 
    }
    \label{fig:prototype}
\end{figure}

\subsubsection{Aggregation-based}
A common solution to calculate prototypes is the aggregation-based method. 

\textbf{Average Pooling. }
In the former few-shot image classification literature, prototypes are computed as the mean of support samples in the embedding space, serving as representative feature vectors for different classes~\cite{snell2017prototypical}. 
When extending this concept to the field of video, a straightforward approach is to average the features of support videos to form a prototype, as shown in Figure~\ref{fig:prototype}(a). 
Although similar strategies have been explored in the literature~\cite{bishay2019tarn, zhang2020ARN}, these methods neglect the intrinsic temporal dimension of video features, leading to inadequate temporal alignment. 
Consequently, several methodologies focus on developing more suitable FSAR samples and prototype matching techniques to fully exploit the spatio-temporal dependencies in videos. 
We illustrate several typical prototype aggregation schemes are illustrated in Figure~\ref{fig:prototype}. 

\textbf{Attention Strategies. }
Attention-based methods are frequently employed to learn prototype representations in few-shot learning, calculating prototypes as weighted sums of few-shot samples, thereby enhancing discriminative capability by focusing on relevant support instances~\cite{gao2019hybrid, yan2022inferring, yang2022sega, liu2022task}. 
Figure~\ref{fig:prototype}(b) illustrates a straightforward method for calculating the prototype with the attention strategy in FSAR, where support samples are combined with a task-dependence learnable weight. 
Several approaches extend this idea to flexibly learn prototypes for novel actions.
For example, Zhu \textit{et al.}~\cite{zhu2021few} find that traditional query-centered loss has limitations in data efficiency and handling outliers or overlapping class distributions. 
The authors present a hybrid attentive prototype-centered learning mechanism aimed at minimizing the adverse effects of outliers and enhancing class separation. 
Additionally, they propose a prototype-centered contrastive learning loss, which involves comparing each prototype against all query-set samples. 
Wang \textit{et al.}~\cite{wang2021temporal} combine the representation of support videos with the similar attention mechanism in~\cite{zhu2021few} to generate high-quality action category prototypes. 
The prototype aggregation adaptive loss is designed to deeply explore the similarity between samples and prototypes, thereby enhancing the identification of inter-class differential details. 
These methods harness the power of attention mechanisms to dynamically weigh and aggregate support samples, thereby enabling more nuanced and adaptive prototype representations, especially in handling diverse and challenging action class distributions.

\textbf{Query-Guided. }
Several methodologies propose learning robust prototype representations with additional knowledge. 
For example, TRX~\cite{perrett2021temporal} constructs action prototypes by considering task-specific information,  as illustrated in Figure~\ref{fig:prototype}(c). 
Specifically, with the guidance of the query video, prototypes are learned by aggregating support samples based on the sub-sequences matching results of query-support pairs.

\textbf{Text-Guided. }
Furthermore, several methods are inspired by the strong generalization ability of textual information to modulate the action prototypes with the guidance of semantic priors to resolve the challenge of inaccurate prototype estimation caused by data scarcity~\cite{ni2024multimodal, xiao2025adaptive}, as illustrated in Figure~\ref{fig:prototype}(d). 
{\color{black}
For instance, Ni \textit{et al.}~\cite{ni2024multimodal} extract semantic features from label texts with a frozen CLIP~\cite{radford2021CLIP} text encoder and further enhanced by a semantic-enhanced module. }
The prototypes are enhanced by semantic features with flexible techniques such as weighted average and multi-head attention. 
Similarly, Wang \textit{et al.}~\cite{wang2023clip} adaptively refine visual prototypes with transferable textual concepts from CLIP, which is achieved by implementing a temporal Transformer to fuse the textual and visual features in the support set. 
These approaches demonstrate how integrating multimodal semantic information can significantly improve the fidelity and effectiveness of prototype representations.

\subsubsection{Learning-based}

Memory networks are frequently employed in few-shot learning to facilitate flexible information storage and retrieval, thereby enhancing generalization performance~\cite{santoro2016meta, cai2018memory, gidaris2018dynamic}, where prototypes are learnable. 
In few-shot action recognition, leveraging this idea needs to consider temporal information in complex video data.

For example, CMN~\cite{zhu2018CMN} maintains a memory containing key-value pairs as the prototypes, where the key memory stores the video representation, and the value memory stores the category label. 
During the inference phase, the memory is updated using the support samples, and recognition results are calculated based on the similarity between the query video features and the features stored in the key memory. 
Additionally, Zhang \textit{et al.}~\cite{zhang2020few} introduce a memory network from a multimodal perspective where prototypes are stored as dynamic visual and textual semantics associated with high-level human activity attributes. 
Qi \textit{et al.}~\cite{qi2020few} introduce an ensemble learning-based architecture, comprising a collection of SlowFast Memory Networks. 
These networks store prototypes at multiple rates, aiming to predict diverse classification results, with the final decision being made by aggregating these results.
Additionally, Liu \textit{et al.}~\cite{liu2022multidimensional} propose a similarity optimization strategy to learn prototypes, aimed at enhancing the ability to identify subtle inter-class differences. 
This strategy consists of two triplets with an adaptive margin to jointly optimize the prototypes.
However, maintaining learnable prototypes introduces additional complexity to the model architecture, potentially increasing computational overhead and training time.

\subsubsection{Compound Methods}

Due to the intertwining of temporal and spatial information in videos, alongside the potential occurrence of actions at diverse speeds, a single approach to prototype calculation may overlook essential video details. 
Consequently, certain methodologies advocate for computing compound prototypes to fortify the resilience of prototype representations. 
For instance, Huang \textit{et al.}~\cite{huang2022compound} summarize each video into compound prototypes consisting of a group of global prototypes and a group of focused prototypes, and then compare video similarity based on the compound prototypes. 
Each global prototype summarizes a specific aspect of the entire video, while focused prototypes target certain timestamps, with bipartite matching applied to handle temporal variations in actions. 
The authors further introduce a method to learn prototypes combined with three types of prototypes~\cite{huang2024matching}. 
The first type describes specific aspects of the action regardless of its timestamp. 
The second type focuses on specific timestamps and uses bipartite matching to handle temporal variations. 
The third type is generated from the timestamp-centered prototypes, regularizing temporal consistency and providing an auxiliary summarization of the entire video. 
Jiang \textit{et al.}~\cite{jiang2024dual} introduce a dual-prototype network that integrates class-specific and query-specific attentive learning. 
The class-specific attentive learning approach enhances prototype representativeness by calculating within-class similarity for each support sample, thereby reducing the influence of noise and outliers. 
In contrast, the query-specific attention mechanism adaptively reweights support samples based on their similarities to each query sample, which increases inter-class discrimination by enlarging differences among support samples. 
Guo \textit{et al.}~\cite{guo2024video} propose a Dual Motion-Guided Attention Learning method to efficiently and effectively identify and correlate motion-related region features from the video level to the task level, allowing the model to construct class prototypes that fully incorporate spatio-temporal relationships. 
{\color{black}
Wu \textit{et al.}~\cite{wu2024efficient} introduce a multi-level matching algorithm that involves comparing visual information and textual information, as well as comparing the visual information conveyed by the support and query samples respectively. 
}
In summary, these methodologies highlight the effectiveness of compound prototype representations in bolstering prototype robustness significantly and enhancing the ability to generalize to novel categories.

\subsubsection{Discussion}

\begin{table*}[t]
\caption{ Results of different category prototype learning methods. }
\label{tab:prototype}
\setlength{\extrarowheight}{0pt}
\setlength{\tabcolsep}{2pt}
\resizebox{\linewidth}{!}{
\begin{tabular}{llllcccccccccc}
\toprule
\multirow{2}{*}{Category}  &\multirow{2}{*}{Method}  & \multirow{2}{*}{Publication} & \multirow{2}{*}{Backbone} & \multicolumn{2}{c}{SSv2-full} & \multicolumn{2}{c}{Kinetics} & \multicolumn{2}{c}{UCF} & \multicolumn{2}{c}{SSv2-small} & \multicolumn{2}{c}{HMDB} \\
&  &  &  & 1-shot  & 5-shot  & 1-shot  & 5-shot  & 1-shot  & 5-shot  & 1-shot  & 5-shot  & 1-shot  & 5-shot  \\ \midrule
\multirow{5}{*}{\begin{tabular}[l]{@{}l@{}}Aggregation-\\ based\end{tabular}} & TRX~\cite{perrett2021temporal}  & CVPR'2021  & ResNet-50  & 42.0 & 64.6 & 63.6 & 85.9 & 78.2 & 96.1 & 36.0 & 56.7 & 53.1 & 75.6 \\
& PAL~\cite{zhu2021few}  & BMVC'2021  & ResNet-50  & 46.0 & 61.1 & 74.6 & 86.6 & -  & -  & -  & -  & -  & -  \\
& TRAPN~\cite{wang2021temporal}  & ACML'2021  & ResNet-50  & -  & -  & 75.1 & 87.0 & 86.6 & 95.9 & -  & -  & 61.3 & 76.8 \\
& MORN~\cite{ni2024multimodal}  & ArXiv'2023 & CLIP-ResNet-50 & -  & 71.7 & -  & 94.6 & -  & 97.7 & -  & -  & -  & 87.1 \\
& CLIP-FSAR~\cite{wang2023clip}  & IJCV'2023  & CLIP-ViT-B  & 61.9 & 72.1 & 89.7 & 95.0 & 96.6 & 99.0 & 54.5 & 61.8 & 75.8 & 87.7 \\ \midrule
\multirow{4}{*}{\begin{tabular}[l]{@{}l@{}}Learning-\\ based\end{tabular}} & CMN~\cite{zhu2018CMN}  & ECCV'2018  & ResNet-50  & -  & -  & 57.3  & 76.0  & -  & -  & 33.4  & 46.5  & -  & -  \\
% Mishra \textit{et al.}\cite{mishra2018generative}  & WACV'2018  & C3D  & -  & -  & -  & -  & -  & 78.7  & -  & -  & -  & 52.6  \\
& Zhang \textit{et al.}~\cite{zhang2020few}  & PR'2020  & 3D ConvNet  & -  & -  & -  & -  & 78.9  & 87.7  & -  & -  & 58.9  & 70.5  \\
& SFMN~\cite{qi2020few}  & ACM MM'2020  & ResNet-50  & -  & -  & 63.7  & 83.1  & -  & -  & -  & -  & -  & -  \\
& Liu \textit{et al.}~\cite{liu2022multidimensional} & TCSVT'2022 & ResNet-50  & 42.1 & 58.4 & 70.2 & 85.3 & 82.0 & 96.4 & -  & -  & 57.3 & 76.8 \\ \midrule
\multirow{5}{*}{\begin{tabular}[l]{@{}l@{}}Compound\\ Methods\end{tabular}} & Huang \textit{et al.}~\cite{huang2022compound}  & ECCV'2022  & ResNet-50  & 49.3 & 66.7 & 73.3 & 86.4 & 71.4 & 91.0 & 38.9 & 61.6 & 60.1 & 77.0 \\
% TSL~\cite{xian2021generalized}  & TPAMI'2022  & R3D  & 20.4  & 31.9  & 86.3  & 92.4  & 89.1  & 95.1  & -  & -  & -  & -  \\
& Huang \textit{et al.}~\cite{huang2024matching}  & IJCV'2024  & ResNet-50  & 52.3 & 67.1 & 74.0 & 86.9 & 74.9 & 92.5 & 42.6 & 61.8 & 61.6 & 77.5 \\
& DPN~\cite{jiang2024dual}  & NeurC'2024 & Resnet-50  & 46.8 & 64.7 & 76.3 & 90.7 & 87.7 & 98.0 & -  & -  & 60.2 & 76.6  \\
& DMGAL-FT~\cite{guo2024video}  & ArXiv'2024 & Resnet-50  & 55.5 & 71.1 & 72.5 & 88.0 & 87.1 & 97.4 & 41.2  & 61.8  & 60.2 & 79.3  \\
& EMP-Net~\cite{wu2024efficient}  & ECCV'2024 & CLIP-ViT-B  & 63.1 & 73.0 & 89.1 & 93.5 & 94.3 & 98.2 & 57.1  & 65.7  & 76.8 & 85.8  \\
\bottomrule
\end{tabular}
}
\end{table*}

The objective of category prototype learning in few-shot action recognition is to learn representative and generalized action representations for novel categories from a few samples, thereby achieving stable recognition performance in each meta-task. 
The conventional average-based method may be affected by sample capacity and extreme samples, making mean-of-class prototypes inadequate in representing the distribution of samples. 
As shown in Table \ref{tab:prototype}, the prototype learning methods designed for FSAR have achieved significant performance.
More recent approaches that enhance class prototype representations with semantic prior knowledge of language image pretrained models have demonstrated significant performance benefits across multiple benchmarks~\cite{ni2024multimodal, wang2023clip}. 
% 

%%%%%%%%%%%%%%%%%%%%%%%%%%%%%%%%%%%%%%%%%%%%%%%%%%%%%%%%%%%%%%%%%
\subsection{Generalized Video Alignment}
The final step in meta-learning-based FSAR is inference.
This stage typically involves comparing query samples to class prototypes and assigning the most likely class based on similarity measures. 
However, due to the temporal nature of video features, alignment is complex.
Several studies perform fine-grained alignment at the frame level rather than treating the video as a whole, emphasizing long-term temporal ordering information to ensure consistency and accuracy of matching.
In addition to temporal information, appearance features, global features, and semantic features contained in the video are also crucial. 
Therefore, several methods propose multi-level alignment to simultaneously focus on different dimensions of video features, thereby achieving more reliable matching. 

\subsubsection{Temporal Alignment}

\begin{figure}
    \centering
    \includegraphics[width=\linewidth]{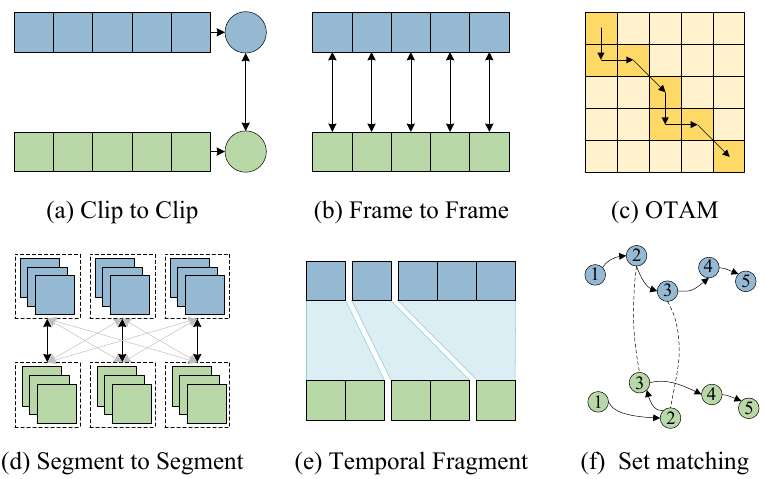}
    \caption
    {
    Typical temporal alignment schemes: 
    (a) Clip-to-Clip Alignment;
    (b) Frame-to-Frame Alignment~\cite{bishay2019tarn};
    (c) OTAM~\cite{cao2020OTAM};
    (d) Segment-to-Segment Alignment~\cite{ perrett2021temporal};
    (e) Temporal Fragment Alignment~\cite{wu2022motion};
    (f) Set Matching~\cite{wang2022hybrid}
    }
    \label{fig:alignment}
\end{figure}

In general few-shot learning, a common alignment approach is to measure the similarity of the query to the learned prototype with distance measures such as cosine or Euclidean distance. 
This approach can be adapted to few-shot action recognition by employing temporal pooling techniques (e.g., average pooling) to derive a clip-wise representation before performing clip-to-clip alignment (Figure~\ref{fig:alignment}(a)), or directly conducting frame-to-frame alignment (Figure~\ref{fig:alignment}(b)). 
For instance, TARN~\cite{bishay2019tarn} implements frame-to-frame alignment and evaluates five distinct distance measures to compute similarity: multiplication, subtraction, neural network, subtraction followed by multiplication and a neural network, as well as Euclidean distance and cosine similarity. 
However, clip-to-clip alignment only captures the co-occurrence rather than the temporal ordering of patterns, which inevitably results in information loss, and frame-to-frame alignment suffers from high computational complexity and sensitivity to noise and temporal misalignment. 

Several methods propose to highlight advancements in temporal alignment for video similarity measurement, balancing strengths in accuracy and adaptability with challenges in computational complexity and parameter tuning. 
For instance, to simplify the sequence matching procedure, ITANet~\cite{zhang2021LIM} learns an implicit temporal alignment that adopts a self-attention mechanism to augment frame features with their temporal context, reducing temporal variation in video instances. 
OTAM~\cite{cao2020OTAM} pioneers the utilization of temporal ordering information by integrating segment distances solely along the ordered temporal alignment path which is illustrated in Figure~\ref{fig:alignment}(c). 
This method extends the Dynamic Time Warping (DTW) algorithm~\cite{muller2007dynamic} to enforce the distance for prediction to preserve temporal ordering. 
Moreover, its differentiable nature enhances flexibility and effectiveness in utilizing long-term temporal information. 
Li \textit{et al.}~\cite{li2021temporal} introduce event boundaries into FSAR to serve as a guiding prior for temporal alignment. 
The authors uniformly sample frames from each action subpart based on these boundaries, rather than considering the entire video. 
This frame sampling strategy is rooted in temporal boundaries and devised to alleviate intra-class variance by ensuring uniform representation across action segments.
Furthermore, a boundary selection module is introduced to pinpoint the initiation and conclusion of actions, facilitating precise alignment of videos with the respective durations. 
TRX~\cite{perrett2021temporal} performs segment-to-segment alignment (Figure~\ref{fig:alignment}(d)), aligning temporally-ordered sub-sequences rather than individual frames to effectively match actions performed at varying speeds and in different parts of videos.
Wang \textit{et al.}~\cite{wang2023molo} develop a motion-augmented long-short contrastive learning method to jointly model the global contextual information and motion dynamics. 
To explicitly integrate the global context into the local matching process, a long-short contrastive objective is applied to enforce frame features to predict the global context of the videos that belong to the same class. 
The authors take frame differences as motion information and design a motion autoencoder to explicitly extract motion features between frame representations by reconstructing pixel motions. 
Li \textit{et al.}~\cite{li2024hierarchical} adopt the Earth Mover’s Distance in the transportation problem to measure the similarity between video features in terms of sub-action representations.  
It computes the optimal matching flows between sub-actions as a distance metric, which is favorable for comparing fine-grained patterns.

Different from either the frame-level alignment~\cite{cao2020OTAM} or the segment-level alignment ~\cite{bishay2019tarn, perrett2021temporal} where the segments are obtained by pre-defined sampling strategies, MTFAN~\cite{wu2022motion} proposes a temporal fragment temporal fragment alignment which is illustrated in Figure~\ref{fig:alignment}(e). 
Inspired by the real-world temporal alignment, which includes ``frame-to-frame", ``segment-to-segment" and ``frame-to-segment" alignment, authors utilize temporal fragments to uniformly represent both frames and segments. 
This approach enables robust matching across videos with different speeds by automatically discovering higher-level segments and performing multi-level temporal fragment alignment. 

The aforementioned methods treat inference in FSAR as either point feature matching or sequence matching. 
In contrast, Wang \textit{et al.}~\cite{wang2022hybrid} propose reformulating distance measure between query and support videos as a set matching problem as illustrated in Figure~\ref{fig:alignment}(f). 
The authors introduce a bidirectional Mean Hausdorff Metric to improve the resilience to misaligned instances which are highly informative and flexible to predict query categories under the few-shot settings. 
Further, they improve the method by devising a temporal set matching metric, which comprises a bidirectional Mean Hausdorff Metric and a temporal coherence regularization, to compute the distances~\cite{wang2024hyrsm++}.

\subsubsection{Multi-level Alignment}
% 
% Temporal alignment poses a significant challenge, prompting several methods to combine multiple alignment strategies in a multi-view approach to enhance classification performance. 

This type of approach advocates integrating multiple temporal alignment methods or considering other critical aspects such as appearance, semantics, or global features of the video along with temporal information for alignment.

Considering multiple temporal alignments enables the model to explore the relationship of videos from different perspectives, resulting in more robust alignments. 
% 
% frame-level and tuple-level
For example, Xing \textit{et al.}~\cite{xing2023boosting} propose a hybrid matching strategy, combining frame-level and tuple-level matching to classify videos with multivariate styles. 
Deng \textit{et al.}~\cite{deng2024two} combine the scores obtained from dynamic alignment and the weighted bipartite graph perfect matching method to derive the final video similarity. 

In addition to temporal alignment, the simultaneous consideration of appearance features is crucial in video matching.
% temporal & spacial
For instance, TA$^2$N~\cite{li2022ta2n} proposes a two-stage strategy to address the misalignment problem from two distinct aspects: action duration and action evolution.
This coarse-to-fine strategy performs a joint spatial-temporal action alignment over videos, to address these two aspects of misalignment sequentially. 
% 
% temporal & spacial
Similarly, Nguyen \textit{et al.}~\cite{nguyen2022inductive} adopt a method that separates appearance and temporal alignments. 
They first compute the appearance similarity score by pairing each frame from one video with its most similar frame from another video, without regard for temporal sequencing. 
Secondly, motivated by the use of temporal order-preserving priors in various video understanding tasks~\cite{su2017order, cao2020OTAM}, the authors also propose a strategy where initial frames from a source video should be mapped to initial frames in a target video, and similarly for subsequent frames. 
This approach encourages the appearance similarity matrix to resemble a temporal order-preserving matrix. The temporal similarity score between the two videos is computed as the negative Kullback-Leibler divergence between these matrices. 
% 
% spatiotemporal
Jiang \textit{et al.}~\cite{jiang2023hitim} combine spatiotemporal self-attention matching and correlated cross-attention matching, where self-attention emphasizes key regions in the spatiotemporal dimension and cross-attention highlights strongly correlated parts between support-set and query-set features. 
% 

% global, temporal, and spatial
Zheng \textit{et al.}~\cite{zheng2022few} propose to consider matching at global, temporal, and spatial levels simultaneously. 
This hierarchical matching method addresses the challenge of escalating complexity in coarse-to-fine matching compounded by the lack of detailed local supervision when focusing on finer-grained visual cues. 
It employs supervised contrastive learning to differentiate videos at various levels and utilizes cycle consistency as weak supervision to align discriminative temporal clips or spatial patches. 
% 
% instance-specific, category-specific, and task-specific
M$^3$Net~\cite{tang2023m3net} integrates various matching functions enabling flexible relation modeling within limited samples to handle multi-scale spatio-temporal variations by leveraging the instance-specific, category-specific, and task-specific matching.

To accommodate speed variance, several methods propose to integrate similarity in sequences of actions at varying speeds. 
% 
% multi-speed
Yu \textit{et al.}~\cite{yu2023multi} generate a multi-speed classification score by integrating the similarities between query videos and support subspaces of varying sampling speeds which is embedding-agnostic and can be combined with most mainstream embedding networks without model re-designs. 
Qu \textit{et al.}~\cite{qu2024mvp} progressively learn and align semantic-related action features at multi-velocity levels to assess the similarity between features from support and query videos with various velocity scales. They subsequently combine all similarity scores in a residual manner.
To prevent multiple velocity features from straying from the underlying motion semantic, the authors incorporate velocity-tailored text information into the video feature via feature interaction across channel and temporal domains at different velocities.

\subsubsection{Discussion}

Achieving effective video alignment in few-shot action recognition is challenging due to the presence of fine-grained motion changes, the temporal nature of videos, and irrelevant sub-movements that can cause misalignment.
To address these issues, existing methods have made significant contributions, as demonstrated by their performance on common benchmarks (Table~\ref{tab:matching}). 
Most current approaches emphasize the acquisition and utilization of temporal information during the alignment, which is crucial for the effectiveness of the matching methods. 
Additionally, the robustness of matching can be further enhanced by integrating multi-level alignment techniques.

\begin{table*}[t]
\caption{Results of different generalized video alignment methods. }
\label{tab:matching}
\setlength{\extrarowheight}{0pt}
\setlength{\tabcolsep}{2pt}
\resizebox{\linewidth}{!}{
\begin{tabular}{llllcccccccccc}
\toprule
\multirow{2}{*}{Category}  & \multirow{2}{*}{Method}  & \multirow{2}{*}{Publication} & \multirow{2}{*}{Backbone} & \multicolumn{2}{c}{SSv2-full}  & \multicolumn{2}{c}{Kinetics}  & \multicolumn{2}{c}{UCF}  & \multicolumn{2}{c}{SSv2-small}  & \multicolumn{2}{c}{HMDB}  \\
&   &  &  & {1-shot} & {5-shot} & {1-shot} & {5-shot} & {1-shot} & {5-shot} & {1-shot} & {5-shot} & {1-shot} & {5-shot} \\ \midrule
\multirow{11}{*}{\begin{tabular}[c]{@{}l@{}}Temporal\\ Alignment\end{tabular}} & TARN~\cite{bishay2019tarn}  & BMVC'2019  & ResNet-50  & -  & -  & 64.8  & 78.5  & -  & -  & -  & -  & -  & -  \\
& OTAM~\cite{cao2020OTAM}  & CVPR'2020  & ResNet-50  & 42.8  & 52.3  & 72.2  & 84.2  & 79.9  & 88.9  & 36.4  & 48.0  & 54.5  & 68.0 \\
& ITANet~\cite{zhang2021LIM}  & IJCAI'2021  & ResNet-50  & 49.2  & 62.3  & 73.6  & 84.3  & -  & -  & 39.8  & 53.7  & -  & -  \\
& TRX~\cite{perrett2021temporal} & CVPR'2021  & ResNet-50  & 42.0  & 64.6  & 63.6  & 85.9  & 78.2  & 96.1  & 36.0  & 56.7  & 53.1  & 75.6 \\
& Cao \textit{et al.}~\cite{cao2021few}  & CVIU'2021  & 3D ConvNet  & -  & -  & -  & -  & 88.7  & 96.8   -  & -  & 63.4  & 79.7 \\
& Li \textit{et al.}~\cite{li2021temporal}  & BMVC'2022  & ResNet-50  & -  & -  & 67.0  & 80.9  &   &   &   &   & 58.6  & 73.8 \\
& MTFAN~\cite{wu2022motion}  & CVPR'2022  & ResNet-50  & 45.7  & 60.4  & 74.6  & 87.4  & 84.8  & 95.1  & -  & -  & 59.0  & 74.6 \\
& HyRSM~\cite{wang2022hybrid}  & CVPR'2022  & ResNet-50  & 54.3  & 69.0  & 73.7  & 86.1  & 83.9  & 94.7  & 40.6  & 56.1  & 60.3  & 76.0 \\
& GgHM~\cite{xing2023boosting}  & ICCV'2023  & ResNet-50  & 54.5  & 69.2  & 74.9  & 87.4  & 85.2  & 96.3  & -  & -  & 61.2  & 76.9 \\
& HyRSM++~\cite{wang2024hyrsm++}  & PR'2023  & ResNet-50  & 55.0  & 69.8  & 74.0  & 86.4  & 85.8  & 95.9  & 42.8  & 58.0  & 61.5  & 76.4 \\
& HCR~\cite{li2024hierarchical}  & CVIU'2023  & R3D  & -  & -  & 75.7  & 86.4  & 88.9  & 95.7  & -  & -  & 67.5  & 79.3 \\ \midrule
\multirow{4}{*}{\begin{tabular}[c]{@{}l@{}}Multi-level\\ Alignment\end{tabular}} & TA$^2$N~\cite{li2022ta2n}  & AAAI'2022  & ResNet-50  & 47.6  & 61.0  & 72.8  & 85.8  & 81.9  & 95.1  & -  & -  & 59.7  & 73.9 \\
& Nguyen \textit{et al.}~\cite{nguyen2022inductive}  & ECCV'2022  & ResNet-50  & 43.8  & 61.1  & 74.3  & 87.4  & 84.9  & 95.9  & -  & -  & 59.6  & 76.9 \\
& HCL~\cite{zheng2022few}  & ECCV'2022  & ResNet-50  & 47.3  & 64.9  & 73.7  & 85.8  & 82.5  & 93.9  & 38.7  & 55.4  & 59.1  & 76.3 \\
& MGCSM~\cite{yu2023multi}  & ACM MM'2023  & ResNet-50  & -  & -  & 74.2  & 88.2  & 86.5  & 97.1  & -  & -  & 61.3  & 79.3 \\
\bottomrule
\end{tabular}
}
\end{table*}

\section{Advanced Topics}
\label{sec:advanced topics}
In this section, we delve into advanced topics in FSAR, extending its applicability to more intricate contexts. 
We review techniques for resolving few-shot action recognition in other modalities or multimodal domains. 
Additionally, we examine methods within unsupervised settings, as well as those addressing cross-domain, incremental, or federated FSAR challenges. 
By comprehensively examining these advanced topics, we aim to deepen our understanding of FSAR and its relevance in solving real-world challenges. 
\subsection{Skeleton-based Learning}

In Section~\ref{sec:methodologies}, 
we summarize the contributions of existing few-shot action recognition works which are primarily based on commonly available RGB modalities. 
However, beyond RGB, human actions can be captured using diverse data modalities that encode unique information, each offering distinct advantages tailored to specific application scenarios.
Skeleton data is a highly effective modality for identifying human behavior, finding extensive applications in traditional behavior recognition tasks. 
Several research efforts aim to address the problem of skeleton identification in few-shot scenarios~\cite{sabater2021one, wang2022temporal, ma2022learning, li2022smam, tseng2022haa4d, memmesheimer2022skeleton, liu2023parallel,  wu2023reconstructed, xu2024enhancing, wei2024novel, lu2024cross, fukushi2024few, yang2024one}. 
Sabater \textit{et al.}~\cite{sabater2021one} introduce skeleton-based representations into FSAR, computing motion representations robust to variable kinematic conditions. This approach involves evaluating query and target motion representations. Specifically, the query action is processed and compared against a target motion sequence to recognize when the anchor action has been performed within the target motion sequence.
Ma \textit{et al.}~\cite{ma2022learning} introduce a spatial matching strategy comprising spatial disentanglement and spatial activation. 
The rationale behind spatial disentanglement stems from the observation that augmenting spatial information for leaf nodes (e.g., the 'hand' joint) enhances representation diversity for skeleton matching. 
To realize spatial disentanglement, the authors advocate for the representation of skeletons in a full-rank space with a rank maximization constraint. 
Furthermore, they propose an attention-based spatial activation mechanism to incorporate disentanglement, whereby the disentangled joints are adaptively adjusted based on matching pairs.
PAINet~\cite{liu2023parallel} integrates two complementary branches aimed at enhancing inter-skeleton and intra-skeleton correlations by leveraging topology and physical information to refine the modeling of interactive parts and joint pairs within both branches.
Moreover, a spatial cross-attention module is utilized to establish joint associations across sequences, while introducing a directional Average Symmetric Surface Metric to identify the closest temporal similarity.
Concurrently, a spatial self-attention mechanism aggregates spatial context within sequences, and a temporal cross-attention network is applied to correct temporal misalignment and calculate similarity. 
% 
% \textit{}
% 
% 
Li \textit{et al.}~\cite{li2022smam} propose a self and mutual adaptive matching module designed to transform these feature maps into more discriminative feature vectors. 
Initially, it exploits both the temporal information inherent in each individual skeleton joint and the spatial relationships among them to extract features. 
Subsequently, the SMAM module dynamically assesses the similarity between labeled and query samples, facilitating feature matching within the query set to differentiate between similar skeletons across various action categories. 
Wei \textit{et al.}~\cite{wei2024novel} propose leveraging a contrastive diffusion graph convolutional network (CD-GCN) for few-shot action recognition. 
Their approach relies on attentional diffusion and contrastive loss to generate robust representations, facilitating the diffusion of diverse graph semantic features.  
Skeleton data inherently provides straightforward yet informative representations of body structure and pose information. 
These skeleton-based methodologies demonstrate robustness against variations in action-independent contexts and have garnered increasing attention.

\subsection{Multimodal Learning}
Because videos have the natural multimodal characteristic, several methods~\cite{xu2018dense, bo2020few,fu2020depth, hong2021video, he2021tbsn, kumar2021few, wang2021audio, majumder2022few, luo2022long, wanyan2023active, Liu2023Lite-MKD} propose to leverage multimodal information (e.g. optical flow, depth, audio and compressed domain data) to enhance the few-shot action recognition procedure. 
For example, AMFAR~\cite{wanyan2023active} proposes to actively find the reliable modality, either RGB or optical flow, for each sample based on task-dependent context information to improve the few-shot reasoning procedure. 
Additionally, optical flow, depth, and appearance features are leveraged in Lite-MKD~\cite{Liu2023Lite-MKD} to train a teacher model to achieve a comprehensive fusion of human movements. 
A student model is utilized to learn to recognize actions from the single RGB modality at a lower computational cost under the guidance of the teacher. 
Mercea \textit{et al.}~\cite{mercea2023text} introduce a unified audio-visual few-shot video classification benchmark. 
They propose a text-to-feature diffusion framework that fuses temporal and audio-visual features through cross-modal attention and generates multi-modal features for novel classes. 
This framework incorporates a hybrid attention mechanism and a diffusion model for multi-modal feature generation, thereby expanding the training dataset with additional novel-class samples. 
{\color{black}
With the emergence of visual language pre-training models, endeavors have been made to leverage the textual information of action labels to enhance the accuracy of few-shot action recognition FSAR~\cite{wang2023clip, ni2024multimodal, cao2024task, wu2024efficient, xing2023multimodal}. }

Effectively integrating other modalities into few-shot action recognition remains a significant challenge. 
Novel methods are required to explore and leverage the correlation between different modalities to improve the robustness and accuracy of the recognition process.

\subsection{Unsupervised Learning}
Unsupervised few-shot action recognition leverages extensive unlabeled video data and bypasses the necessity for base-class labels or supervised pertaining. 
The primary challenge lies in effectively adapting unsupervised representations to few-shot tasks without explicit supervision, as unsupervised video representations often lack specificity for novel classes.
% , leading to suboptimal performance in the few-shot scenario. 
% 
MetaUVFS~\cite{patravali2021unsupervised} is introduced as the first unsupervised meta-learning algorithm for few-shot action recognition which requires no base-class labels or supervised pre-trained backbones, needing only a single training session to perform competitively, often surpassing supervised approaches.
It utilizes over 550K unlabeled videos from Kinetics700~\cite{carreira2019kinetics-700} to train a two-stream 2D and 3D CNN architecture~\cite{he2016resnet, hara2018resnet3d} via contrastive learning to capture both appearance-specific spatial features and action-specific spatio-temporal features in videos. 
The core of MetaUVFS is an Action-Appearance Aligned Meta-adaptation module, which focuses on action-oriented video features in relation to appearance features through few-shot episodic meta-learning over unsupervised hard-mined episodes. 
The two streams of features are combined by learning a spatio-temporal alignment of appearance over action features. 
This strategy enhances intra-class similarity and reduces inter-class confusion in few-shot learning scenarios.

Additionally, HyRSM++~\cite{wang2024hyrsm++} extends the FSAR method to unsupervised tasks which adopts the idea of ``clustering first and then meta-learning” paradigm to construct few-shot tasks and exploits unlabeled data for training and achieve impressive performance. 

Unsupervised few-shot action recognition needs neither base-class labels nor a supervised pretrained backbone, enabling the generalization of novel actions while reducing the dependency on costly supervised pertaining.

\subsection{Cross-Domain}
Current few-shot action recognition methods assume abundant labeled videos of base classes and few-shot videos of novel classes originating from the same domain. 
However, this assumption is impractical as novel categories often emerge from different data domains with diverse spatial and temporal characteristics in real-world applications~\cite{gao2020pairwise, samarasinghe2023cdfsl, markham2024understanding, guo2025dmsd, wang2024tamt}. 
To tackle this issue, Gao \textit{et al.}~\cite{gao2020pairwise} propose investigating cross-domain few-shot action recognition. 
They employ an attentive adversarial network to align spatiotemporal dynamics, addressing domain discrepancies and enhancing the discrimination of learned domain-invariant features through a pairwise margin discrimination loss.
CDFSL-V~\cite{samarasinghe2023cdfsl} utilizes self-supervised learning and curriculum learning to balance information from both the source and target domains, thereby learning generic features for cross-domain few-shot action recognition. 
Li \textit{et al.}~\cite{li2024learning} introduce a method that identifies and leverages transferable knowledge from a model pretrained on a large image dataset. 
During adaptation, this approach preserves transferable temporal dynamics while updating the image encoder.
Additionally, SSAlign~\cite{xu2023augmenting} and RelaMix~\cite{peng2023exploring} explore few-shot domain adaptation strategies that diverge from the conventional definition of FSAR proposed in this paper but inspire further investigation into adapting models across domains.
Cross-domain few-shot action recognition addresses the challenge of recognizing actions in novel domains with limited labeled data, which is crucial for applications requiring the adaptation of existing models to new environments or scenarios.

\subsection{Incremental Learning}

Incremental learning enables models to continuously adapt to new data without requiring complete retraining, which is crucial for addressing evolving environments~\cite{castro2018end, wu2019large}. 
Incremental few-shot learning specifically targets the incremental acquisition of novel concepts with limited samples~\cite{tao2020few, perez2020incremental}. 
In the context of few-shot action recognition, Feng \textit{et al.}~\cite{feng2024spatiotemporal} propose to resolve the incremental few-shot action recognition challenge by employing a spatiotemporal attention routing mechanism and an orthogonal projection capsule layer. 
The spatiotemporal attention routing mechanism encodes spatial and temporal transformation information and explores part-whole relationships of actions to mitigate catastrophic forgetting. 
The orthogonal projection capsule layer is designed to maintain a sufficient distance between the prototypes of old and novel classes, thereby avoiding overfitting by considering spatial-temporal features. 

\subsection{Federated Learning}

With the widespread use of action recognition technology, ensuring data privacy and addressing ethical concerns have become increasingly critical. 
This involves making the decision-making process traceable, thereby bolstering user trust in the technology. 
In few-shot action recognition, this necessitates the development of privacy-preserving methods such as differential privacy and federated learning to enable effective action recognition while safeguarding personal data.
Federated Learning (FL) is a promising solution for many computer vision applications due to its effectiveness in handling data privacy and communication overhead. 
For example, Tu \textit{et al.}~\cite{tu2024fedfslar} propose investigating the federated few-shot learning problem, which involves collaboratively learning a classification model from multiple FL clients to recognize unseen actions with only a few labeled video samples.

\section{Future work}
\label{sec:future work}

\subsection{More Realistic Dataset}
Current research on few-shot action recognition has primarily focused on a few standard datasets (i.e., UCF~\cite{soomro2012ucf101}, HMDB~\cite{kuehne2011hmdb}, Kinetics~\cite{carreira2017kinetics}, Something to Something~\cite{goyal2017something} and EPIC-Kitchens~\cite{damen2018EPIC-Kitchens}), typically comprising predefined categories of actions within specific contexts. 
However, real-world action data is often characterized by increased complexity, diversity, and noise, surpassing the coverage provided by existing datasets. 
Addressing this gap requires efforts toward constructing datasets with broader categories and finer-grained actions. For example, commonplace actions like ``walking" can be further categorized into variations such as ``walking slowly", ``walking fast" and ``walking at a steady pace". 
Furthermore, gathering action data across diverse scenes, including indoor and outdoor environments, various weather conditions, and different lighting conditions, is crucial for enhancing model robustness.
% 
% Dataset design should also consider the long-tail distribution of action classes to ensure sustained model performance with a limited number of examples per class. 
Integrating real-world noise is also essential. 
Introducing background interference and noise, such as non-target actions, occlusion, camera shake, etc., effectively simulates challenges encountered in practical application environments. These efforts collectively contribute to creating more comprehensive datasets and facilitate the development of robust few-shot action recognition models.

{\color{black}
\subsection{More Realistic Settings}
Currently, FSAR research predominantly follows the traditional N-way K-shot FSL setting. 
However, as discussed in Sec 3.1.2, there is growing interest in more general FSL settings. 
Therefore, it is imperative to expand FSAR into more realistic settings in future research. 

Besides traditional few-shot learning tasks, significant future research avenues encompass generalized FSAR, open-set FSAR, open-world FSAR, and so on. 
Generalized FSAR requires the model not only to classify video samples from novel action categories accurately but also from seen categories~\cite{xian2021generalized}. 
With the advancement of large models, pre-trained video models are now capable of covering an increasing variety of action categories. 
This necessitates that the model rapidly generalizes to novel categories based on only a few videos, while not conflating them with a large number of seen categories.
Open-set FSAR requires the model to recognize few-shot categories and reject samples that do not belong to either seen or few-shot categories. 
Differences between similar action classes may be subtle and temporal-dependent, thus heightening the intricacy of model recognition. 
This presents new challenges for the model to accurately comprehend action concepts.
Additionally, open-world FSAR necessitates precise identification of actions in query videos in settings marked by scant training data and unreliable labels. 
Robustness is crucial for the model, which needs to accurately comprehend action concepts and remain unaffected by poor data quality. 
This is particularly significant in the present era of pre-training models, as the quantity and quality of samples utilized for pre-training are not consistently assured. 

Inspired by the open-vocabulary action recognition task, an important research scenario for FSAR encompasses the recognition of not only the novel class (represented by support samples) to which the query sample belongs but also the specific action label. 
This is a highly promising and challenging task, demanding that the model accurately learn novel class concepts from few-shot samples but also exhibit strong semantic knowledge generalization capabilities. 
Currently, there is limited study on these FSAR settings that hold practical significance, thereby presenting \textbf{another important research direction} with promising applications. 
These emerging research directions will expand the applicability of FSAR, facilitating its effective implementation across a broader spectrum of real-world scenarios. 

}

% \subsection{Exploring Novel Modalities}
\subsection{Modality Extension}

Early few-shot action recognition research primarily utilizes RGB videos due to their widespread availability and accessibility. 
As discussed in Section~\ref{sec:advanced topics}, several approaches have been developed to leverage skeleton, audio, depth, and optical flow modalities, as well as multimodal combinations, to address the FSAR task. 
These additional modalities can offer unique advantages for FSAR, enabling more accurate and robust recognition in various application scenarios. 
However, the increasing demand for recognizing daily life activities necessitates the use of additional modalities such as infrared sequences, point clouds, event streams, acceleration, radar, and WiFi. 
For instance, infrared sequences can capture clear action information in low-light conditions, and point clouds provide precise three-dimensional spatial structures. 
Future research should focus on developing effective methods for extracting features from these unexplored modalities and designing appropriate matching techniques to enhance FSAR performance. 
Moreover, the world is inherently multimodal, making it crucial to explore multimodal few-shot action recognition. 
However, the varying levels of noise and quality across different modalities complicate the integration process. 
Future methods need to effectively align and synchronize information across these modalities to enable realistic applications.

\subsection{Adoption of Large Language Model}

% advantage:
% 1. abundant prior knowledge
% 2. representation learning capabilities
% 3. process and integrate information in various modalities
% 
% 
Large language models (e.g., GPT-4~\cite{achiam2023gpt}, LLAVA~\cite{liu2023llava}, Deepseek-R1~\cite{guo2025deepseek}, etc.) and multimodal large models (e.g., LLAMA~\cite{touvron2023llama}, Qwen-VL~\cite{wang2024qwen2}) have demonstrated remarkable performance in areas such as natural language processing and image generation. 
These models have been extensively pre-trained to simulate the abundant prior knowledge that humans possess before learning novel concepts with few-shot samples. 
This presents a valuable research opportunity to achieve better few-shot action recognition.
Additionally, the representation learning capabilities of large models allow them to capture higher-level features and abstract representations of actions. 
In few-shot learning, leveraging pre-trained large models has the potential to facilitate faster adaptation to novel action categories and improve generalization in few-shot action recognition.
Moreover, the ability of multimodal large models to effectively process and integrate information in various modalities has the potential to significantly enhance the accuracy and robustness of FSAR models, especially in complex environments with diverse action variations. 
Current applications~\cite{wang2023clip, ni2024multimodal, pei2023d} mainly aim to directly leverage semantic prior knowledge learned by the language-image pretrained model to obtain more generalizable representations for videos. 
Applying large models to few-shot action recognition still holds significant promise and presents under-explored challenges.

% fine-tuning and deployment
There are several issues to consider in the process of using large models. 
Firstly, the fine-tuning and deployment for large models are challenging. 
Consequently, effective model adaptation strategies are necessary to reduce computational overhead and memory requirements. 
Leveraging techniques such as adaptors, network compression, and dynamic architecture tuning can enhance efficiency without compromising performance.
% 
% dataset biases and domain shifts
Additionally, because current large models are not pre-trained on the target domain or even on video data, addressing dataset biases and domain shifts is critical. 
Future research should focus on source-free domain generalization methods tailored to ensure robust performance across diverse real-world scenarios. 
It is also essential to prevent catastrophic forgetting to maintain generalization ability. 
{\color{black}
Moreover, In-Context Learning (ICL) is a popular algorithm for few-shot learning especially for LLMs, still has insufficient practical application in the context of FSAR. 
Given the potential computational challenges that large models may encounter when processing several video data together, the adaptation of ICL methods to such models presents a challenging problem. }
Although large visual-language models have shown outstanding results in multiple multimodal tasks, utilizing them to effectively process multimodal video data in FSAR remains a challenge. 

{\color{black}
Recent efforts integrating Vision-Language Models (VLMs) into FSAR have predominantly relied on the CLIP model~\cite{wang2023clip, chen2023video, kahatapitiya2024victr}. 
However, there is a lack of endeavors in FSAR utilizing video large models~\cite{zhang2023video} that perform remarkable multimodal understanding and reasoning abilities. 
    (i) The FSAR task involves the processing of video data, which typically incurs significant computational overhead. 
    This challenge is particularly pronounced when employing VLMs. 
    (ii) Given the continual emergence of novel action categories, the adaptation to these changes represents a crucial challenge, despite the rich common-sense knowledge possessed by VLM. 
    (iii) Furthermore, in the context of fine-grained FSAR, it is notably challenging for Vision-and-Language Models (VLM) to capture subtle temporal differences in certain actions.
In the future, it may be necessary to consider the utilization of Video Large Models from the perspective of balancing computational complexity and performance, fine-grained action recognition, and interpretability of large model decisions, as well as the design of in-context learning algorithms. }

In addition to visual and semantic information, many other modalities need to be considered in FSAR, such as audio and sensor signals. 
The variations across different modalities make it very complex to adjust and synchronize information across these different modalities. 
This encourages research to develop large models that can integrate diverse modalities for FSAR.

\section{Conclusion}
\label{sec:conclusion}

In this paper, we provide a comprehensive survey of few-shot action recognition. 
Beginning with a foundational discussion, we define key concepts, formulate the problem, and introduce relevant datasets and evaluation metrics. 
Then, we review existing methods, categorizing them into generative-based and meta-learning where meta-learning-based methods have three primary types: learning video instance representations, learning category prototypes, and generalized video alignment. 
We explore recent advancements in few-shot action recognition, highlighting emerging trends and innovative techniques. 
We summarize the challenges and identify critical research directions, offering insights into unresolved questions and opportunities for future research.
This survey aims to inform researchers and practitioners about the current landscape of few-shot action recognition, while also laying the groundwork for future advancements and collaborations within the academic community. 

\vspace{5mm}
\noindent
\textbf{Data Availability:} 
This survey introduces commonly used datasets for FSAR, summarized in Section~\ref{sec:benchmarks}. 
These publicly available datasets include HMDB (\url{https://serre-lab.clps.brown.edu/resource/hmdb-a-largehuman-motion-database}), UCF101 (\url{https://www.crcv.ucf.edu/data/UCF101.php}), Kinetics (\url{https://github.com/cvdfoundation/kinetics-dataset}), SSv2 (\url{https://20bn.com/datasets/something-something}), and EPIC-Kitchens (\url{https://epic-kitchens.github.io/2021}).
% 
% \ref{sec:benchmarks}

    % CMN~\cite{zhu2018CMN}
    % TARN~\cite{bishay2019tarn}
    % OTAM~\cite{cao2020OTAM}
    % PAL~\cite{zhu2021few}
    % MTFAN~\cite{wu2022motion}
    % GgHM~\cite{xing2023boosting}
    % CLIP-FSAR~\cite{wang2023clip}

% \clearpage
% \bibliographystyle{ieeetr}
% \bibliography{ref}
% \bibliography{sn-bibliography}

{\small
\bibliographystyle{abbrvnat}
\bibliography{ref}% common bib file
}

\end{document}